\documentclass[twoside]{article}
\usepackage{graphicx}

\usepackage[accepted]{aistats2024}
\usepackage{xcolor}

\def\P{{\mathbb P}}

\def\Tcal{{\mathcal T}}

\def\Xcal{{\mathcal{X}}}
\def\Ncal{{\mathcal{N}}}
\def\Fcal{{\mathcal{F}}}

\def\Dcal{{\mathcal D}}
\def\Scal{{\mathcal S}}

\def\Acal{{\mathcal A}}
\def\Hcal{{\mathcal H}}
\def\Xcal{{\mathcal X}}

\def\Mcal{{\mathcal M}}
\def\Qcal{{\mathcal Q}}
\def\Ncal{{\mathcal N}}
\def\Ocal{{\mathcal O}}

\def \chisq {{\chi^2_\Qcal}}
\def \holder {{H\"{o}lder }}

\usepackage{hyperref}
\usepackage{amsmath}
\usepackage{amssymb}
\usepackage{amsthm}
\usepackage[OT1]{fontenc}
\usepackage{subcaption} 
\usepackage{caption}

\usepackage{graphicx}
\usepackage{bm}
\usepackage{algorithm}
\usepackage{algorithmic}
\usepackage{natbib}

\newtheorem{theorem}{Theorem}[section]

\newtheorem{lemma}[theorem]{Lemma}
\newtheorem{definition}[theorem]{Definition}
\newtheorem{assumption}[theorem]{Assumption}

\newcommand{\cG}{\mathcal{G}}

\newcommand{\cD}{\mathcal{D}}
\newcommand{\cP}{\mathcal{P}}
\newcommand{\cT}{\mathcal{T}}
\newcommand{\cH}{\mathcal{H}}
\newcommand{\cA}{\mathcal{A}}
\newcommand{\cF}{\mathcal{F}}

\usepackage{amsmath,amsfonts,bm}

\def\eqref#1{equation~\ref{#1}}

\def\1{\bm{1}}

\DeclareMathAlphabet{\mathsfit}{\encodingdefault}{\sfdefault}{m}{sl}
\SetMathAlphabet{\mathsfit}{bold}{\encodingdefault}{\sfdefault}{bx}{n}

\newcommand{\E}{\mathbb{E}}

\newcommand{\R}{\mathbb{R}}

\DeclareMathOperator*{\argmax}{arg\,max}

\usepackage{hyperref}
\usepackage{url}
\usepackage{titlesec}
\titlespacing{\section}{0pt}{3pt}{3pt}
\titlespacing{\subsection}{0pt}{2pt}{2pt}
\titlespacing{\paragraph}{0pt}{2pt}{2pt}

\begin{document}

\twocolumn[
\aistatstitle{Policy Evaluation for Reinforcement Learning from Human Feedback: A Sample Complexity Analysis}

\aistatsauthor{Zihao Li \And Xiang Ji \And  Minshuo Chen \And Mengdi Wang }

\aistatsaddress{ Princeton University \And  Princeton University \And Princeton University \And Princeton University } ]

\begin{abstract}
A recently popular approach to solving reinforcement learning is with data from human preferences. In fact, human preference data are now used with classic reinforcement learning algorithms such as actor-critic methods, which involve evaluating an intermediate policy over a reward learned from human preference data with distribution shift, known as off-policy evaluation (OPE). Such algorithm includes (i) learning reward function from human preference dataset, and (ii) learning expected cumulative reward of a target policy. Despite the huge empirical success, existing OPE methods with preference data often lack theoretical understandings and rely heavily on heuristics. In this paper, we study the sample efficiency of OPE with human preference and establish a statistical guarantee for it. Specifically, we approach OPE with learning the value function by fitted-Q-evaluation with a deep neural network. By appropriately selecting the size of a ReLU network, we show that one can leverage any low-dimensional manifold structure in the Markov decision process and obtain a sample-efficient estimator without suffering from the curse of high data ambient dimensionality. Under the assumption of high reward smoothness, our results \textit{almost align with the classical OPE results with observable reward data}. To the best of our knowledge, this is the first result that establishes a \textit{provably efficient} guarantee for off-policy evaluation with RLHF. 
\end{abstract}

\allowdisplaybreaks

\section{Introduction}

\textit{Reinforcement Learning with Human Feedback} (RLHF) is a popular approach to solve reinforcement learning problems using data based on human feedback. In recent years, RLHF has achieved significant success in large language models, clinical trials, auto-driving, robotics, etc. \citep{ouyang2022training, gao2022scaling,glaese2022improving,hussein2017imitation,jain2013learning,kupcsik2018learning,menick2022teaching,nakano2021webgpt,novoseller2020dueling,chakraborty2024parl}. In conventional reinforcement learning, the goal of the learner is to find the optimal policy using data containing reward observations; in RLHF, the learner does not have direct access to the reward signal but instead can only receive human preferences over actions or trajectories, which are assumed to be made in line with the actual reward. RLHF approaches are popular because it is believed that humans are accurate at comparing and ranking a small number of items based on their preference, and such preferences can be modeled with classic human choice models such as the Plackett-Luce (PL) ranking model \citep{Bradley1952RankAO}. 

One popular way to do RLHF is by first constructing a reward function based on human preference data and then using this reward in a classic reinforcement learning algorithm \citep{christiano2017deep,ibarz2018reward}. A prominent example is RLHF with actor-critic methods, which are commonly used in robotics applications \citep{lee2021pebble,liu2022meta,liang2022reward}. In such methods, the actor is responsible for learning and improving the policy of the agent so that the agent's policy can converge to the optimal policy; meanwhile, the critic evaluates the value function of the current policy in an off-policy manner in order to guide the actor to improve the current policy in the following iteration. In such an algorithm, the intermediate policy that the critic needs to evaluate could be arbitrary and non-optimal, and the data it uses during the evaluation are generated by the policy from previous steps, which is different from the one to be evaluated. Such evaluation is called off-policy, and it is crucial to study its error in order to understand the algorithm's convergence. 


In fact, off-policy evaluation (OPE) can be formulated more generally beyond the actor-critic setting and is itself an important problem in reinforcement learning. When on-policy Monte Carlo rollout is time-consuming, costly, or risky, OPE becomes necessary and has therefore found many applications such as healthcare \citep{murphy2001marginal}, recommendation system \citep{chapelle2011empirical}, education \citep{mandel2014offline}, robotics \citep{levine2020offline} and dialogue system \citep{jiang2021towards}. Due to the mismatch between the visitation distributions of the behavior and target policies, evaluation in the off-policy setting is entirely different from the on-policy one and can be much more challenging and delicate. Most existing studies of OPE focus on the standard setting and rely on direct reward observations in the dataset. In contrast, OPE with preference data has been barely studied.

There has been a recent line of work that theoretically studies RLHF in the offline setting. For learning the optimal policy, \citet{zhu2023principled} proposed the pessimistic MLE algorithm for stepwise preferences and analyzed its suboptimality in the linear function approximation case. \citet{zhan2023provable} studied the same problem under trajectory-wise preferences in the general function approximation setting. \citet{li2023reinforcement} studied offline RLHF under the dynamic choice model. However, these works all focus on the policy learning problem, and there has yet been any result that studies RLHF in the unrealizable setting, where the function approximation cannot perfectly express the reward and value function and has to incur an approximation error. In this paper, we study the estimation error of OPE with human preference data. To stay close to the practice, we consider the function approximation with deep neural networks. We seek to delineate how things like the distribution mismatch and network size can affect the final estimation error. Our challenges are three-folded: 
    (i) The reward itself is not directly observable and needs to be learned from the human preference data;
    (ii) our reward learning is subject to error from distribution mismatch and unrealizable function approximation. This error is propagated to the later policy evaluation and needs to be carefully controlled. 
    (iii) In many applications, we need to handle a high-dimensional state-action space, in which deep neural networks might need a lot of samples to operate.


\paragraph{Our results.} In this work, we address these challenges by analyzing the {F}itted {Q}-{E}valuation with {H}uman {P}references  algorithm, which is based on Maximum Likelihood Estimation and Fitted Q-evaluation. Specifically, our contributions are as follows: 
\begin{itemize}
    
    \item We provide an upper bound for the estimation error of OPE with human preferences and show how distribution shift, intrinsic dimension, underlying model class and sample size of the human preference data can affect the OPE error. Our analysis shows that the error from the reward learning part using human preferences propagates only in a moderate way and does not bottleneck the final estimation error. 
    
    \item We consider function approximation with deep neural networks for nonparametric reward and Q-functions. In this setting, we prove that by choosing hyperparameters appropriately, deep ReLU network can adapt to the intrinsic low-dimension structure of the preference data and keep the estimation error mostly independent from the the ambient dimension. 
\end{itemize}

Notably, our result matches the existing error bounds for standard OPE and demonstrates OPE with human preferences can be as sample-efficient, even in the absence of an observable reward. This provides a theoretical support for the actor-critic methods with RLHF in practice \citep{lee2021pebble,liang2022reward}, which essentially solve OPE with human preference data. To the best of our knowledge, this paper offer the first theoretical guarantee for off-policy evaluation with human preference.

\subsection{Related Work}
\paragraph{RLHF for policy learning.}  In recent years RLHF and inverse reinforcement learning (IRL) has been widely applied to robotics, recommendation system, and large language model \citep{ouyang2022training, lindner2022interactively,menick2022teaching,jaques2020human,lee2021pebble,nakano2021webgpt,chandrasekaran2021human}. Policy optimization applied to train text generation systems with pre-specified rewards \cite{cho2018coherent} and ranking systems \cite{zhang2023unified}. To model human choice behavior, \citep{shah2015estimation,ouyang2022training, saha2022efficient} learn reward from pairwise comparison and ranking.  \cite{pacchiano2021dueling} study pairwise comparison with function approximation in pairwise comparison. To learn the optimal policy, most of them implement actor-critic-type algorithms such as PPO or SAC. From a theoretical view,
\citet{zhu2023principled,zhan2020humanguided} study various cases of preference-based comparison in trajectory bandit problems with linear function approximation and establish a theoretical guarantee for their algorithm. \cite{li2023reinforcement} study RLHF under human feedback setting. However, none of these theoretical works have studied policy evaluation or a deep neural network setting.
\paragraph{Off-policy evaluation.} 
Except for the FQE studied in this paper, there have been various methods to do off-policy evaluation. One popular type is using importance sampling to reweigh samples by the distribution shift ratio \citep{precup2000eligibility}, but importance sampling suffers from large variance, especially when the sampling policy is unknown, which causes an error with an exponential term in the length of
the horizon in the worst case. To address this issue, some variants with reduced variance such as
marginal importance sampling (MIS) \citep{xie2019towards} and doubly robust estimation \citep{jiang2016doubly, thomas2016data,uehara2020off} have been developed. For the tabular setting with complete data coverage, \citep{yin2020asymptotically} show that MIS is an asymptotically
efficient OPE estimator, which matches the Cramer-Rao lower bound in \cite{jiang2016doubly}. \cite{duan2020minimaxoptimal} studies OPE with linear function approximation. \cite{zhang2022FQE} studies OPE with differentiable function approximation. Moreover, a line of
work \citet{zhan2022offline, lee2021optidice,nachum2019dualdice,nachum2020reinforcement} focuses on policy evaluation using MIS and
linear programming, without function approximation.


\paragraph{Deep neural network in reinforcement learning.}In recent years, deep learning has enjoyed tremendous empirical success in RL applications\citep{mnih2013playing,silver2016mastering,gu2016continuous,mao2016resource}. In RLHF, deep neural
networks are also crucial when the state-action space is immense or continuous, such as \citet{ouyang2022training,stiennon2020learning,wu2021recursively}. Specifically, the popular GPT 3.5 is fine-tuned by first learning reward function from the human feedback dataset, and then implementing PPO to learn the optimal policy. Recently, there have been some theoretical studies of deep RL.  \cite{fan2020theoretical} focuses on the
online policy learning problem and studies DQN with feed-forward ReLU network; \cite{yin2022offline} studies offline policy learning with realizable, general differentiable function approximation. Notably, a recent study \cite{ji2022sample} provides an analysis of the estimation error of nonparametric FQE
using a feed-forward CNN network. However, all these works study deep RL with an observable reward given by the environment, which lies in the standard RL regime. To the best of our knowledge, there has not been any result on the
theoretical guarantee one can achieve with deep RL when the data are human preferences.

\paragraph{Notation}
For a scalar $a>0,\lceil a\rceil$ denotes the ceiling function, which gives the smallest integer that is no less than $a ;\lfloor a\rfloor$ denotes the floor function, which gives the largest integer that is no larger than $a$. For any scalars $a$ and $b, a \vee b$ denotes $\max (a, b)$ and $a \wedge b$ denotes $\min (a, b)$. For a vector or a matrix, $\|\cdot\|_0$ denotes the number of nonzero entries and $\|\cdot\|_{\infty}$ denotes the maximum magnitude of entries. Given a function $f: \mathbb{R}^D \rightarrow \mathbb{R}$ and a multi-index $s=\left[s_1, \cdots, s_D\right]^{\top}, \partial^s f$ denotes $\frac{\partial^{|s|} f}{\partial x_1^{s_1} \cdots \partial x_D^{s_D}} \cdot\|f\|_{L^p}$ denote the $L^p$ norm of function $f$. For an arbitrary space $\Xcal$, we use $\Delta(\Xcal)$ to denote the set of all probability distribution on $\Xcal$. For two vectors $x,y\in \R^d$, we denote $x\cdot y = \sum_i^d x_iy_i$ as the inner product of $x,y$. We denote the set of all probability measures on $\Xcal$ as $\Delta(\Xcal)$. We use $[n]$ to represent the set of integers from $0$ to $n-1$.  For every set $\Mcal\subset\Xcal$ for metric space $\Xcal$, we define its $\epsilon$-covering number with respect to norm  $\|\cdot\|$ by $N(\Mcal, \|\cdot\|,\epsilon)$.We adopt the convention $0 / 0=0$. Given distributions $p$ and $q$, if $p$ is absolutely continuous with respect to $q$, the Pearson $\chi^2$-divergence is defined as $\chi^2(p, q):=\mathbb{E}_q\big[\big(\frac{\mathrm{d} p}{\mathrm{~d} q}-1\big)^2\big]$.

\section{Preliminaries}
\subsection{Markov Decision Process}

We define a finite-horizon MDP model $M = (\Scal,\Acal, H, \{P_h\}_{h\in[H]},\{r_h\}_{h\in[H]} )$, $H$ is the horizon length,  in each step $h\in[H]$ , the agent starts from state $s_h$ in the state space $\Scal$, chooses an action  $a_h \in \Acal$ with probability $\pi_{h}(a_h\mid s_h)$ , receives a reward of $r_h(s_h,a_h)$ and transits to the next state $s'$ with probability $P_h(s'\mid s_h,a_h)$. Here $\Acal$ is a finite action set with $|\Acal|$ actions and $P_h(\cdot| s_h,a_h) \in \Delta(s_h,a_h)$ is the transition kernel condition on state action pair $(s, a)$. For convenience we assume that $r_h(s,a) \in [0,1]$ for all $(s,a, h) \in \Scal \times \Acal\times[H]$. Without loss of generality, we assume that the initial state of each episode $s_0$ is fixed. Note that this will not add difficulty to our analysis. For any policy $\pi = \{\pi_h\}_{h\in[H]}$ the state value function is $$
V_h^\pi(s) = \E_\pi\bigg[\sum_{t = h}^H r_t(s_t, a_t)\bigm\vert s_h = s\bigg],
$$
and the action value function is
$$
Q_h^\pi(s,a) = \E_\pi\bigg[\sum_{t = h}^H r_t(s_t, a_t)\bigm\vert s_h = s,a_h = a\bigg],
$$
here the expectation $\E_\pi$  is taken
with respect to the randomness of the trajectory induced by
$\pi$, i.e. is obtained by taking action $a_t \sim \pi_t(\cdot\mid s_t)$ and observing $s_{t+1} \sim P_h(\cdot\mid s_t,a_t)$. For any function $f: \Scal\rightarrow \R$, we define the transition operator: \begin{align*}
    &\cP_h^\pi f(s,a)\\
    &\qquad= \E[f(s_{h+1},a_{h+1})\pi(a_{h+1}\mid s_{h+1})\mid s_h =s,a_h =a].
\end{align*}

We also denote the Bellman operator at time $h$ under policy $\pi$ as $\cT_h^\pi$: $$\cT_h^\pi f (s,a) := r_h(s,a) + \cP_h^\pi f(s,a). $$
For a given policy $\pi = \{\pi_h\}_{h\in[H]}$, we use $q_h^\pi(s,a)$ to denote the state-acton distribution induced by $\pi$, i.e. the probability of visiting $(s,a)$ on step $h$ when the policy is $\pi$.

\subsection{Riemannian Manifold}
Real-world data often exhibit low-dimensional geometric structures and can be viewed
as samples near a low-dimensional manifold. In this section, we briefly introduce the concept of Riemannian manifolds. Let $\mathcal{M}$ be a $d$-dimensional Riemannian manifold isometrically embedded in $\mathbb{R}^D$. A chart for $\mathcal{M}$ is a pair $(U, \phi)$ such that $U \subset \mathcal{M}$ is open and $\phi: U \rightarrow \mathbb{R}^d$ is a homeomorphism, i.e., $\phi$ is a bijection, its inverse and itself are continuous. Two charts $(U, \phi)$ and $(V, \psi)$ are called $\mathcal{C}^k$ compatible if and only if
\begin{align*}
&\phi \circ \psi^{-1}: \psi(U \cap V) \rightarrow \phi(U \cap V)\\
 \text { and } \quad &\psi \circ \phi^{-1}: \phi(U \cap V) \rightarrow \psi(U \cap V)
\end{align*}
are both $\mathcal{C}^k$ functions ( $k$-th order continuously differentiable). A $\mathcal{C}^k$ atlas of $\mathcal{M}$ is a collection of $\mathcal{C}^k$ compatible charts $\left\{\left(U_i, \phi_i\right)\right\}$ such that $\bigcup_i U_i=\mathcal{M}$. An atlas of $\mathcal{M}$ contains an open cover of $\mathcal{M}$ and mappings from each open cover to $\mathbb{R}^d$. We call a manifold \textit{smooth} if it has a $\mathcal{C}^{\infty}$ atlas.

We also define the reach \citep{niyogi2006finding,federer1959curvature} of a manifold to characterize the curvature of $\mathcal{M}$.
\begin{definition}[Reach, Definition 2.1 in \cite{aamari2019estimating}]
    The medial axis of $\mathcal{M}$ is defined as $\overline{\mathcal{T}}(\mathcal{M})$, which is the closure of
\begin{align*}
    \mathcal{T}(\mathcal{M})=&\bigg\{x \in \mathbb{R}^D \mid \exists x_1 \neq x_2 \in \mathcal{M} \text{ such that }\\
    &\left\|x-x_1\right\|_2=\left\|x-x_2\right\|_2=\inf _{y \in \mathcal{M}}\|x-y\|_2\bigg\} .
\end{align*}

The reach $\omega$ of $\mathcal{M}$ is the minimum distance between $\mathcal{M}$ and $\overline{\mathcal{T}}(\mathcal{M})$, i.e.
$$
\omega=\inf _{x \in \overline{\mathcal{T}}(\mathcal{M}), y \in \mathcal{M}}\|x-y\|_2 .
$$
\end{definition} 
Roughly speaking, reach measures how fast a manifold "bends" in the embedded Euclidean space. A manifold with a large reach "bends" relatively slowly. On the contrary, a small $\omega$ signifies more complicated local geometric structures, which are possibly hard to fully capture.

\subsection{\holder Functions on Manifolds}

To capture the smoothness of the Bellman operator and the human preference reward function, we define a \holder function on a smooth manifold $\Mcal$. 
\begin{definition}[\holder function]\label{def-holder}
 Let $\mathcal{M}$ be a $d$-dimensional compact Riemannian manifold isometrically embedded in $\mathbb{R}^D$. Let $\left\{\left(U_i, {P}_i\right)\right\}_{i \in \mathcal{A}}$ be an atlas of $\mathcal{M}$ where the ${P}_i$ 's are orthogonal projections onto tangent spaces. For a positive integer $k$ and $l \in(0,1]$ such that $k+l = \alpha$, a function $f: \mathcal{M} \mapsto \mathbb{R}$ is $\alpha$-\holder continuous if for each chart $\left(U_i, {P}_i\right)$ in the atlas, we have\begin{itemize}
     \item[(i)]$f \circ {P}_i^{-1} \in C^s$ with $\left|D^\mathbf{k}\left(f \circ {P}_i^{-1}\right)\right| \leq 1$ for any $|\mathbf{k}| \leq k, x \in U_i$;
     \item[(ii)] for any $|\mathbf{k}|=k$ and $x_1, x_2 \in U_i$,
\begin{align*}
&\left|D^{\mathbf{k}}\left(f \circ {P}_i^{-1}\right)\right|_{{P}_i\left(x_1\right)}-\left.D^{\mathbf{k}}\left(f \circ {P}_i^{-1}\right)\right|_{{P}_i\left(x_2\right)} \mid\\
&\quad\leq\left\|{P}_i\left(x_1\right)-{P}_i\left(x_2\right)\right\|_2^l .
\end{align*}

 \end{itemize}

Moreover, we denote the collection of $\alpha$-\holder functions on $\mathcal{M}$ as $\mathcal{H}^{\alpha}(\mathcal{M})$. For a multivariate function, we say it is $\alpha$-\holder if it is $\alpha$-\holder in every dimension.

\end{definition}
Definition \ref{def-holder} requires that all $s$-th order derivatives of $f \circ {P}_i^{-1}$ are Hölder continuous. We recover the standard Hölder class on an Euclidean space if ${P}_i$ is the identity mapping. 

\subsection{ReLU Neural Network Class}

We consider a  rectified linear unit
(ReLU) neural network as our function class. Given an input $x\in \R^D$, an $L$-layer ReLU neural network computes an output as \begin{equation}\label{eq:def-relu}
f(x) = W_L \cdot\operatorname{ReLU}(W_{L-1}\cdots\operatorname{ReLU}(W_1x+b_1)\cdots+b_{L-1})+b_L,
\end{equation}
where $W_1,\dots,W_L$ and $b_1,\dots,b_L$ are weight matrices and vectors of proper sizes. Here ReLU$(x) = \max \{x, 0\}$ .

We denote $\mathcal{E}(R,\tau ,L,p,I)$ as a class of neural networks with bounded weight parameters and bounded output: \begin{align}\label{eq:nn-def}
    &\mathcal{E}(R, \tau, L, p, I)=\bigg\{f \mid f({x})\text{ in the form of \eqref{eq:def-relu}} \notag\\
    &\text{ with $L$-layers and width bounded by $p$, and satisfies} \notag\\
&\|f\|_{\infty} \leq R,\left\|W_i\right\|_{\infty, \infty} \leq \tau,\left\|{b}_i\right\|_{\infty} \leq \tau \text { for } i=1, \ldots, L, \notag\\ &\sum_{i=1}^L\left\|W_i\right\|_0+\left\|{b}_i\right\|_0 \leq I\bigg\}.
\end{align}

\section{Preference-Based Fitted Q-Evaluation}
We consider the \textit{preference-based}
nonparametric off-policy evaluation (OPE) problem in a finite-horizon time-inhomogeneous MDP. The transition kernel $\{P_h\}_{h\in[H]}$ and reward function $\{r_h\}_{h\in[H]}$ are both unknown to us. Specifically, our objective is to evaluate the value of $\pi$ from a fixed initial distribution $\xi$ over horizon $H$, given by \begin{align*}
    v^\pi := \E\bigg[\sum_{h=1}^H r_h(s_h,a_h)\bigm\vert s_1 \sim \xi\bigg],
\end{align*}
where $a_h \sim \pi(\cdot | s_h)$ and $s_{h+1} \sim P_h(s_{h+1}| s_h, a_h)$. We are given two datasets for every timestep $h\in[H]$. To learn the transition information in the MDP, we are given a dataset containing samples of state transitions $\Dcal_h := \{(s_{h,k}, a_{h,k}, s_{h,k}')\}_{k=1}^K$, which is i.i.d. sampled by an unknown behavior policy $\pi_0$. We also assume a given dataset containing human preference, $\Dcal_h^{HF}:= \{(\tilde{s}_{h,k}^{(1)}, \tilde{a}_{h,k}^{(1)}), (\tilde{s}_{h,k}^{(2)}, \tilde{a}_{h,k}^{(2)}), y_{h,k}\}_{k=1}^{K_{HF}}$, in which for each $h\in [H]$, $\{(\tilde{s}_{h,k}^{(1)}, \tilde{a}_{h,k}^{(1)}), (\tilde{s}_{h,k}^{(2)}, \tilde{a}_{h,k}^{(2)})\}_{k=1}^{K_{HF}}$ are i.i.d. samples generated from a fixed distribution $\eta_h$, which could come from a pretrained model \citep{ouyang2022training} or a reward-free exploration algorithm \citep{lee2021pebble}, $y_{h,k}$ is the human preference made at episode $k$ that will be clarified in the following.  We assume that human preference among state-action pairs is generated under the following model.
\begin{assumption}[Human preference model]\label{ass:rew-model}
    For every $h\in[H]$, given  arbitrary $M$ state-action pairs $\{s_h^{(m)}, a_h^{(m)}\}_{m\in[M]}$, we assume that the human prefers $y_h $  according to the following model: 
    \begin{align}\label{eq:pref-def}
        &\P\big(y_h = (s_h^{(i)}, a_h^{(i)}) \mid (s_h^{(m)}, a_h^{(m)})_{m\in[M]}\big)\propto \exp({r_h(s_h^{(i)}, a_h^{(i)})}),
    \end{align}
    where $i\in[M]$.
\end{assumption}
Such assumption has been widely adopted to model human choice behavior in RL with human preference, e.g. \cite{sharma2017inverse,wulfmeier2016maximum,ziebart2008maximum,zhou2020learning, ouyang2022training, zhu2023principled}. Note that under Assumption \ref{ass:rew-model}, shifting the reward function by a constant for all $(s,a)$ simultaneously doesn't affect human preference probability, thus reward function is not identifiable. For identifiability of the reward,  without loss of generality, we also need to make the following assumption.
\begin{assumption}[Reward identifiability]\label{ass:rew-iden}
    We assume that there exists a fixed state-action pair $(s_h^{f}, a_h^{f})$ for every $h\in[H]$, such that $r_h(s^{f}, a^{f}) = 0$. Moreover, we assume that the human agent can prefer $(s^f,a^f)$ for all $(h,k)\in[H]\times[K]$.
\end{assumption}
With Assumption \ref{ass:rew-iden}, we further assume that $y_{h,k}$, the human preference at step $k$, is chosen among $\{(\tilde{s}_{h,k}^{(1)}, \tilde{a}_{h,k}^{(1)}), (\tilde{s}_{h,k}^{(2)}, \tilde{a}_{h,k}^{(2)}), (s^f, a^f)\}$, generated independently according to Assumption \ref{ass:rew-model} with $M=3$. Assumption \ref{ass:rew-iden} eliminates arbitrary constant shifting in the reward so that we can be determined by a unique reward function from Assumption \ref{ass:rew-model}. Such assumption is akin to empirical works such as \cite{christiano2017deep,stiennon2020learning}, in which $(s^f,a^f)$ is interpreted as human rater making an error, and also widely accepted by choice models literature, in which a fixed option is assumed to be of zero-utility, e.g. \cite{adusumilli2019temporal,buchholz2021semiparametric}.  
We are now ready to present our algorithm. 
\begin{algorithm}[H]
   \caption{FQE with human preferences}
   \begin{algorithmic}[1]\label{alg-main}
   \REQUIRE  Datasets $\{\cD_h^{HF}\}_{h\in[H]}, \{\cD_h\}_{h\in[H]}$,  neural network classes $\mathcal{F}(R, \tau, L, p, I)$ and $\mathcal{G}(\Tilde{R}, \tilde{\tau}, \tilde{L}, \tilde{p}, \tilde{I})$ defined by \eqref{eq:nn-def}, target policy $\pi = \{\pi_h\}_{h\in[H]}$
   	 \FOR{step $h=H, \dots,1$}
   	 \STATE Learn  $\bar{r}_h$ with MLE :
\begin{align*}
    {r}^l_h &= \arg\max_{g\in\cG_h}\sum_{k=1}^{K}\bigg\{f(y_{h,k}) \\
    -&\log\bigg(\exp(f(s^f,a^f)+\sum_{i=1}^2\exp\big(f(s_{h,k}^{(i)},a_{h,k}^{(i)})\big)) \bigg)\bigg\}.
\end{align*}
\STATE Normalize the reward across states and set $
\widehat{r}_h(s,a) = {r}^l_h(s,a) - {r}^l_h(s^{f},a^{f}).
$\label{step-align}
\STATE Update $\widehat{Q}^\pi_h = \widehat{\Tcal}_h^\pi \widehat{Q}^\pi_{h+1}$ by \eqref{eq:FQE-update}.

   	 \ENDFOR\\
     \STATE Compute policy value with the estimated Q-function:
\begin{equation*}
    \widehat{v}^\pi = \int_\Xcal \widehat{Q}_1^\pi(s,a)\xi(s)\pi_1(a ~|~ s) d(s,a).
\end{equation*}
    \STATE \textbf{Output:} $\widehat{v}^\pi$
   \end{algorithmic}
\end{algorithm}
The pseudocode for our algorithm is presented in Algorithm \ref{alg-main}. For every $h\in[H]$, our algorithm consists of two stages:
    (i) We learn the reward by a \textit{maximum likelihood estimation} method. Specifically, we use a deep ReLU function class $\mathcal{G}(\Tilde{R}, \tilde{\tau}, \tilde{L}, \tilde{p}, \tilde{I})$  to do the function approximation for all $h\in[H]$.  By Assumption \ref{ass:rew-iden}, we align the estimated reward across states by Step  \ref{step-align} to ensure that $\widehat{r}_h(s^f,a^{f}) = 0$. Our estimation for reward aligns with the line of RLHF empirical studies including \citet{ouyang2022training, lee2021pebble,stiennon2020learning,wu2021recursively}, in which a reward function is learned from a human preference dataset assuming an underlying softmax human policy; (ii)
To estimate $v^\pi$
, we use \textit{neural FQE} to estimate $Q_h^\pi$
in a backward, recursive fashion, with $\cF(R,\tau,L,p,I)$ as the neural network class for all $h\in[H]$. $\widehat{Q}_h^\pi$, our estimate at step
$h$, is set to be $\widehat{T}_h^\pi \widehat{Q}_{h+1}^\pi$, whose update rule is based on the Bellman equation: \begin{align}\label{eq:FQE-update}
\mathcal{\widehat T}_h^\pi \widehat Q_{h+1}^\pi= &\arg\min_{f\in\cF} \sum_{k=1}^{K}\bigg(f(s_{h,k},a_{h,k}) - \widehat{r}_h(s_{h,k},a_{h,k})\nonumber\\
&\quad- \int_\cA \widehat{Q}_{h+1}^\pi(s_{h,k}',a)\pi_h(a ~|~ s_{h,k}') d a\bigg)^2,\end{align} 
where $\widehat{\Tcal}_h^\pi$
is an intermediate estimate of the Bellman operator $\Tcal_h^\pi$, $\widehat{Q}_{h+1}^\pi$ is an intermediate estimate of $Q^\pi_{h+1}$. 

\section{Main Results}

In this section, we prove an upper bound on the estimation error of $v^\pi$ by Algorithm \ref{alg-main}. First, let us
state two assumptions on the MDP.

\begin{assumption}[Low-dimensional state-action space]\label{ass-low-dim-sa}
    The state-action space $\mathcal{X}$ is a $d$-dimensional closed Riemannian manifold, which is embedded isometrically in Euclidean space ${R}^D$. We also assume that there exists constant $B>0$ such that $\|x\|_{\infty} \leq B$ for any $x \in \mathcal{X}$. The reach of $\mathcal{X}$ is $\omega>0$.
\end{assumption}
Assumption \ref{ass-low-dim-sa} characterizes the low intrinsic dimension of the MDP embedded in high dimensions. We consider the case that the “intrinsic dimension” of $\Xcal$ is $d \ll D$. Such a setting is common in practice because the representation or features people use are often redundant compared to the latent structures of the problem. For instance, images of a dynamical system are widely believed to admit such low-dimensional latent structures  \citet{gong2019intrinsic, hinton2006reducing, pope2021intrinsic,osher2017low}. Common pre-trained language models can also have a low intrinsic dimension as shown in recent empirical studies \citep{aghajanyan2020intrinsic,hu2021lora}.

\begin{assumption}[Bellman completeness]\label{ass:smthness}
Under target policy $\pi$, for any time step $h$ and any $f\in\Fcal$, we have $\cP^\pi_h f \in \cH^{\alpha}(\mathcal{M})$, where $0<\alpha\leq+\infty$.

\end{assumption}
Assumption \ref{ass:smthness} is about the closure of a function class under the Bellman operator, which is widely adopted in RL literature \citet{ji2022sample,chen2019informationtheoretic, yang2020function, du2020good}. In fact, some classic MDP settings also have this property implicitly, e.g., linear MDP \citep{jin2020provably} or RKHS MDP \citep{yang2020function}. Such an assumption is necessary for the Bellman residual: without such an assumption, \cite{wang2020statistical} provides a lower bound that the sample complexity is exponential in the horizon for guaranteeing a constant error in OPE. 
\begin{assumption}\label{ass:rew-smooth}
     We assume that the human preference reward function can be expressed as $r_h(s,a) = f_h(\psi_h(s,a))$ for every $h\in[H]$, where $\psi_h: \Xcal \rightarrow [0, B']^{\tilde{d}}$ and $f: [0, B']^{\tilde{d}} \rightarrow [0,1]$. Furthermore, $\psi_h \in \cH^{\tilde{\alpha}}(\Xcal)$ and $f\in\cH^{\tilde{\alpha}}([0,B']^{\tilde{d}})$, where $0<\tilde{\alpha}\leq +\infty$.
\end{assumption}
\begin{figure}[htp]
    \centering
    \includegraphics[width=8cm]{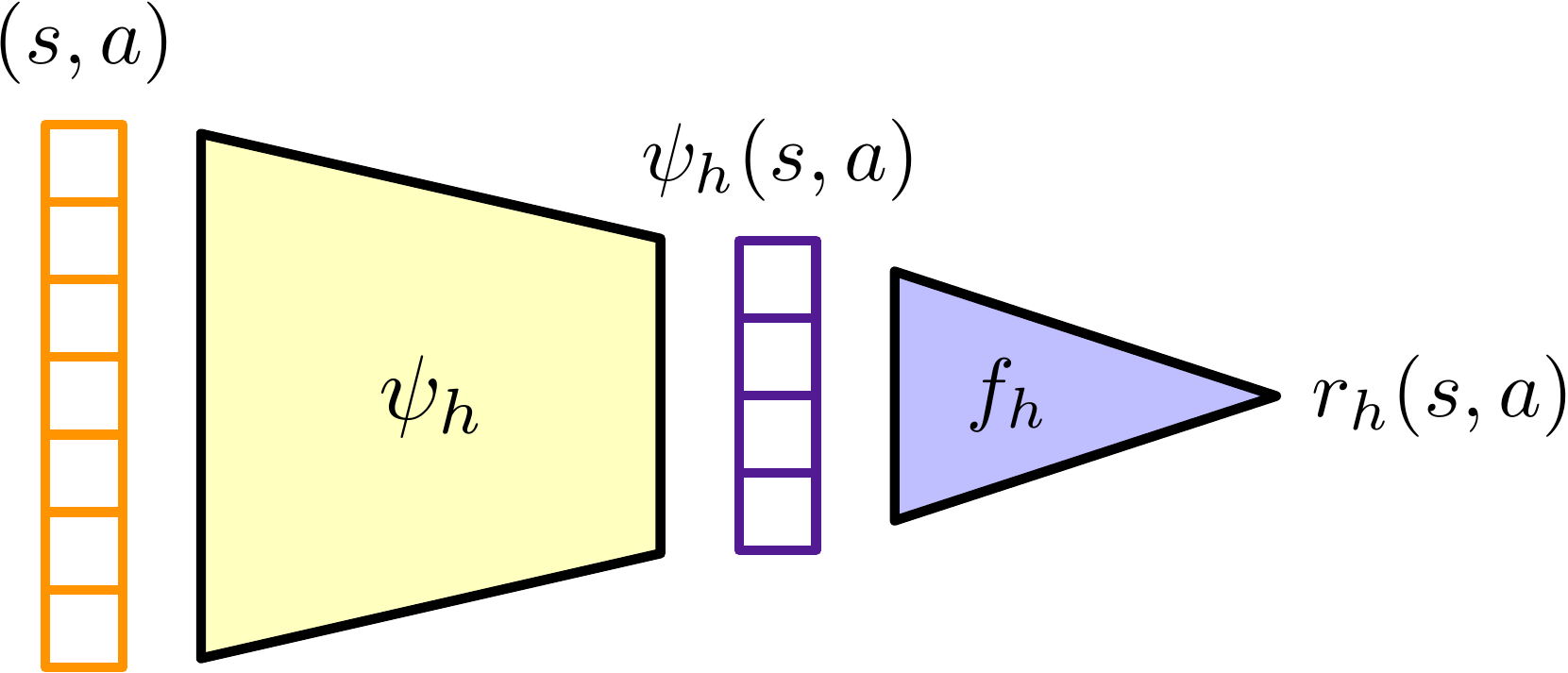}
    \caption{Reward function with low-dim feature.}
    \label{fig:galaxy}
\end{figure}
In Assumption \ref{ass:rew-smooth}, we assume that the reward function admits an underlying feature $\psi_h$, which is sufficiently smooth and maps the input to a space with lower dimension $\tilde{d}$. Empirically, many studies have shown that learning human preferences requires only a smaller dataset and architecture, which implies that such preference rewards admit a low dimension latent feature, see e.g. \citet{christiano2017deep,wang2022skill,stiennon2020learning,chandrasekaran2021human}.   We summarize our results for reward learning in Theorem \ref{thm-reward-estimate}:
\begin{theorem}[Reward convergence rate]\label{thm-reward-estimate}
    With Assumption \ref{ass:rew-smooth}, for every $0<\epsilon_0<1$, if the ReLU class $\mathcal{G}_h(\Tilde{R}, \tilde{\tau}, \tilde{L}, \tilde{p}, \tilde{I})$ satisfies \begin{align*}
    &\tilde{L}=O\bigg(\log \frac{1}{\epsilon_0}\bigg), \tilde{p} =O\bigg(\epsilon_0^{-\frac{2\tilde{d}}{\tilde{\alpha}}}\cdot \tilde{d}\bigg),\tilde{R} = 1, \\
    &\tilde{I} =O\bigg(\epsilon_0^{-\frac{2\tilde{d}}{\tilde{\alpha}}}\cdot \tilde{d}\cdot\log \frac{1}{\epsilon_0}\bigg),\tilde{\tau} = \max\{B, 1, \sqrt{\tilde{d}}, \omega^2\}, 
    \end{align*}
    then we have \begin{align}\label{eq:rew-eps}
        &\E\bigg[\int_\Xcal \big(\widehat{r}_{h}(s,a) - r_h(s,a) \big)^2 d\eta_h(s,a)\bigg]\nonumber\\
        &\quad\leq O\bigg(\epsilon_0^2 +\epsilon_0^{-2\tilde{d}/\tilde{\alpha}} \cdot\frac{\log K_{HF}}{K_{HF}}\cdot\log^3\frac{2}{\epsilon_0}\bigg)  
    \end{align}
    for all $h \in [H]$. Here the expectation $\E[\cdot]$ is taken over the data-generating process of $\Dcal^{HF}$. Recall that $\eta_h$ is the sampling distribution of $(s,a)$ for the human preference data.
\end{theorem}
\begin{proof}
    See Appendix \ref{sec:proof-main} for details.
\end{proof}
Theorem \ref{thm-reward-estimate} upper bounds the mean square error given by neural MLE. Specifically, the estimation error consists of two terms: (i) the complexity of function class $\cG(\tilde{R}, \tilde{\tau},\tilde{L},\tilde{p},\tilde{I})$, measured by its covering number, and (ii)  model approximation error $\epsilon_0$. Such a scenario is the classic bias-variance tradeoff, which needs to be carefully balanced by selecting the appropriate network size. Specifically, by setting $\epsilon_0 = O\big(K_{HF}^{-\frac{\tilde{\alpha}}{2\tilde{\alpha}+2\tilde{d}}}\big)$, we have the mean square error bounded by $O\big(K_{HF}^{-\frac{2\tilde{\alpha}}{2\tilde{\alpha}+2\tilde{d}}}\cdot \log^3 K_{HF}\big)$. 
To depict the distribution shift of the dataset, we introduce the following metric: \begin{definition}[Restricted $\chi^2$-divergence]
    For two distribution $q_1, q_2$ defined on a sample space $\Xcal$, and a function class $\Qcal$ of functions that maps $\Xcal$ to $\R$,  we define their restricted $\chi^2$ divergence  $\chi_{\mathcal{Q}}^2\left(q_1, q_2\right)$  as
$$
\chi_{\mathcal{Q}}^2\left(q_1, q_2\right)=\sup _{f \in \mathcal{Q}} \frac{\mathbb{E}_{q_1}[f]^2}{\mathbb{E}_{q_2}\left[f^2\right]}-1.
$$
\end{definition}
We are now ready to present our main theorem, which provides a statistical upper bound for the estimation error in off-policy evaluation. 
\begin{theorem}\label{thm-main-fqe}
    Suppose Assumption \ref{ass-low-dim-sa} and \ref{ass:smthness} hold. By Algorithm \ref{alg-main}, we have \begin{align*}
\E[|\widehat{v}^\pi - v^\pi|] \leq O\Bigg(&H\kappa_1 \cdot \bigg(K^{-\frac{\alpha}{2\alpha +d}} + \sqrt{\frac{D}{K}}\bigg)\cdot\log^{3/2}K\\
&+H(\kappa_1+\kappa_2) \cdot K_{HF}^{-\frac{\tilde{\alpha}}{2\tilde{\alpha}+2\tilde{d}}}\cdot \log K^{3/2}\Bigg)
    \end{align*}
    if ReLU network class $\mathcal{G}_h(\Tilde{R}, \tilde{\tau}, \tilde{L}, \tilde{p}, \tilde{I})$ satisfies 
    \begin{align*}
    &\tilde{L}=O\bigg( \frac{\tilde{\alpha}}{2\tilde{\alpha} +\tilde{d}}\cdot\log K_{HF}\bigg), \tilde{p}  =O\bigg(K_{HF}^{\frac{\tilde{d}}{2\tilde{\alpha} +\tilde{d}}}\cdot \tilde{d}\bigg), \\
    &\tilde{I} =O\bigg(\frac{\alpha}{2\alpha +d}\cdot K_{HF}^{\frac{\tilde{d}}{2\tilde{\alpha}+\tilde{d}}} \cdot \tilde{d}\cdot\log K_{HF}\bigg),\\
    &\tilde{\tau} = \max\{B, 1, \sqrt{\tilde{d}}, \omega^2\}, \tilde{R} = 1, \end{align*} 
and $\cF_h(R, \tau, L, p, I)$ satisfies \begin{align*}
        &L = O\bigg( \frac{\alpha}{2\alpha +d}\cdot\log K\bigg), p =O\bigg(K^{\frac{d}{2\alpha +d}}\bigg),\\
        &I = O\bigg(\frac{\alpha}{2\alpha +d}\cdot K^{\frac{d}{2\alpha +d}}\cdot\log K\bigg),\\
        & \tau = \max\{B,1,\sqrt{d},\omega^2\}, R =H.
    \end{align*}
 
The distributional mismatch of the transition dataset is captured by $\kappa_1 = \sum_{h=1}^H \sqrt{1+\chi_\Qcal^2(q_h^\pi, q_h^{\pi_0})}$, and the distributional mismatch of the human preference dataset is captured by $\kappa_2 = \sum_{h=1}^H \sqrt{1+\chi_\Qcal^2(q_h^\pi, \eta_h)}$, in which $q_h^\pi$ and $q_h^{\pi_0}$
are the visitation distributions of $\pi$ and $\pi_0$ at step $h $ respectively, $\eta_h$ is the sampling distribution of $\big\{(s_h^{(i)}, a_h^{(i)})\big\}_{i=1}^2$  and $\Qcal$ is
the Minkowski sum between the ReLU function class $\cF(R, \tau, L, p, I)$ and the \holder function class $\Hcal^{\alpha}(\Xcal)$. 
 $O(\cdot)$ hides factors depending on $\log D,\log H, B, B', \omega$ and surface area of $\Xcal$.
\end{theorem}
\begin{proof}
    See Appendix \ref{sec:rew-proof} for details.
\end{proof}
Theorem \ref{thm-main-fqe} explains the success of fine-tuning processes in RLHF such as \citet{ouyang2022training, stiennon2020learning}, in which a compact dataset with human feedback and a relatively small model is adequate for acquiring a precise reward function. Particularly, the reward learning error contribute a $O(H(\kappa_1+ \kappa_2) \cdot K_{HF}^{-\frac{\tilde{\alpha}}{2\tilde{\alpha}+2\tilde{d}}}\cdot \log K_{HF}^{3/2})$ term in the total estimation error of $v^\pi$. When $\tilde{d} \ll d$ and $\tilde{\alpha} \gg \alpha$, i.e. the reward function has a smaller intrinsic dimension and higher smoothness, the error inherited from reward learning is much smaller than the error caused by learning the underlying MDP. 
We further discuss the implication of Theorem \ref{thm-main-fqe} in details: 
\paragraph{Provable benefits of human feedback.} Designing a satisfying reward function is hard in reinforcement learning. When learning from human feedback, we have no access to a given reward function as in previous off-policy evaluation works \citep{ji2022sample,duan2020minimaxoptimal,uehara2022review}.  Compared to the guarantee standard neural FQE such as \cite{ji2022sample}, in which the rewards are explicitly given by the environment,  Algorithm \ref{alg-main} learns reward function from human feedback by neural MLE, which results in an additional $\tilde{O}\big(K_{HF}^{-\frac{\tilde{\alpha}}{2\tilde{\alpha}+2\tilde{d}}}\big)$ term. Such a term does not affect our error bound as long as $K_{HF} \asymp K^{\frac{\alpha/\tilde{\alpha}}{(2\alpha +d)/ (2\tilde{\alpha}+2\tilde{d})}}$.  We would like to highlight that when the human-preference reward has an intrinsic smooth and low dimensional structure, i.e. $\tilde{\alpha} \gg \alpha$  and $\tilde{d}\ll d$, we can train the reward with a $\Dcal^{HF}$ with size $K_{HF}\ll K$. This explains why a small dataset is sufficient for fine-tuning a large model. Under such a circumstance, our evaluation error is $\tilde{O}(K^{-\frac{\alpha}{d+ 2\alpha}})$, which matches the standard results in doing FQE with an observable reward, such as \cite{ji2022sample}. Although OPE with human preference contains less information as the reward is unobservable in the sampled dataset, Theorem \ref{thm-main-fqe} shows that it is as efficient as standard OPE with observable reward function. 
\paragraph{(II) Adaptation to intrinsic dimension. } Note that our estimation error is mainly influenced by the intrinsic dimensionality $d$ and $\tilde{d}$, instead of the representation dimensionality $D$. Consequently, the bound of estimation error for our method is much smaller than the error in methods that ignore the problem's intrinsic dimension, such as \cite{nguyentang2022sample}.
\paragraph{(III) Charaterizing distribution shift.} The term $\kappa_1$ and $\kappa_2$ in our result quantify the divergence between the visitation distribution of the target policy and the sampling policy of human preference data and transition information, using a restricted $\chi^2$-divergence. It's worth noting that this restricted $\chi^2$-divergence is always less than or equal to the widely used absolute density ratio \citep{nguyentang2022sample,chen2019information,xie2020q}, and often, it can be significantly smaller. This occurs because the probability measures $q_h^\pi$ and $q_h^{\pi_0}$ may vary considerably in small regions of the sample space, while their integration of a smooth function in $\mathcal{Q}$ across the entire sample space can be close. The absolute density ratio captures the former variation, whereas the restricted $\chi^2$-divergence captures the latter. We also highlight that although the error in the reward learning part propagates to the FQE part, the distribution mismatch in human preference only exhibits an additive relation with the distribution mismatch of the transition dataset ${\Dcal}$ without amplification.
\label{headings} 
\section{Proof Sketch}
In this section we briefly sketch the proofs of efficiencies of Algorithm \ref{alg-main}. For detailed proof, see Appendix \ref{sec:rew-proof} and \ref{sec:proof-main}.
\paragraph{Proof outline for Theorem \ref{thm-reward-estimate}.} In the reward estimation stage, by implementing MLE on human preference probability $\rho(\cdot\mid s, a^{(0)}, a^{(1)}, a^{default})$,  we claim that with high probability, the model error under the distribution of the sampling policy can be bounded by the approximation error of neural network class $\cG_h$ and its complexity. Specifically, we prove that \begin{align*}
&\E\bigg[\E_{\eta_h}\big[\|{\rho}_{l,h}(\cdot\mid (s_h^{(0)},a_h^{(0)}), (s_h^{(1)},a_h^{(1)}))\\
&\qquad- \rho_{b,h}(\cdot\mid (s_h^{(0)},a_h^{(0)}), (s_h^{(1)},a_h^{(1)}))\|_1^2\big]\bigg] \\
&\qquad\leq O\bigg(\frac{1}{K_{HF}} \log\big(H \cdot N(\cG_h, \|\cdot\|_\infty, 1/K_{HF})\big) + \epsilon\bigg),\nonumber
\end{align*}
where $\rho^l_h$ is the preference distribution learned by MLE, $\epsilon$ is the approximation error of neural function class $\cG_h$. Notably, our method takes model misspecification into account, which is novel in MLE error analysis. By the definition of preference in \eqref{eq:pref-def} and the condtion that $r_h(s^f,a^f) = 0$, we can prove that \begin{align*}
    &\E\bigg[\int_{\Xcal} |r_h(s,a) - \widehat{r}_h(s,a) |^2 d\eta_h(s,a) \bigg]\\
    &\qquad\leq O\bigg(\frac{1}{K_{HF}} \log\big(H \cdot N(\cG_h, \|\cdot\|_\infty, 1/K_{HF})\big) + \epsilon\bigg) .
\end{align*}
Finally, by the approximation ability of neural networks, we can give an upper bound for the function class complexity in terms of $O(\epsilon^{-d/\alpha}\cdot\log 1/\epsilon)$, while maintaining a given approximation error. Set $\epsilon = \epsilon_0^2$ and we conclude the proof.

\paragraph{Proof outline for Theorem \ref{thm-main-fqe}.} Theorem \ref{thm-main-fqe} is proved by an inductive fashion. In particular, with notation \begin{align*}
    &E_h := \E\bigg[\int_\Xcal \big| \Tcal_h^\pi Q_{h+1}^\pi(s,a) - \widehat{\Tcal}_h^\pi\widehat{Q}_{h+1}^\pi(s,a)\big|dq_h^\pi(s,a)\bigg],\\
    & \bar{T}_h f(s,a) = \widehat{r}_h(s,a) + \cP_h f(s,a),
\end{align*}

by Cauchy-Schwarz inequality we can prove that \begin{align*}
    &E_h \leq E_{h+1}\\
    &+ \sqrt{1+ \chi_\Qcal^2(q_h^\pi, q_h^{\pi_0})}\\
    &\quad\cdot \E\bigg[{\int_\Xcal \big(\widehat{\Tcal}_h^\pi \widehat{Q}_{h+1}^\pi(s,a) - \bar{\Tcal}_h^\pi \widehat{Q}_{h+1}^\pi(s,a) \big)^2dq^{\pi_0}_{h}(s,a)}\bigg]^{1/2}\\
    &+ \sqrt{1+ \chi_\Qcal^2(q_h^\pi, \eta_h)}\\
    &\quad\cdot \E\bigg[{\int_\Xcal \big( {r}_{h+1}(s,a) - \widehat{r}_{h+1}(s,a) \ \big)^2d\eta_h(s,a)}\bigg]^{1/2}
\end{align*}
which consists of a distribution shift term,  an $L^2$ norm between difference of $\widehat{\Tcal}_h$ and $\bar{\Tcal}_h$ projected on the estimate of $Q_{h+1}$, and an estimation error for human preference reward function. We further decomposed the right-hand side into the generalization and approximation error in neural network regression, and the error inherited in reward estimation. Note that we face the trade-off between the size of neural net size and its approximation limit. We recommend the readers to Appendix \ref{sec:proof-main} for details. By choosing the neural network class size appropriately, we prove that
\begin{align*}
    &\E\bigg[{\int_\Xcal \big(\Tcal_h^\pi \widehat{Q}_{h+1}^\pi(s,a) - \widehat{\Tcal}_h^\pi \widehat{Q}_{h+1}^\pi(s,a) \big)^2dq^{\pi_0}_{h}(s,a)}\bigg]\\
    &\qquad\lesssim H\bigg(K^{-\frac{2\alpha}{2\alpha +d}}+ K_{HF}^{-\frac{2\tilde{\alpha}}{2\tilde{\alpha}+2\tilde{d}}}+{\frac{D}{K}}\bigg).
\end{align*}
Sum everything together and we conclude the proof.

\section{Conclusion}

This paper studies nonparametric off-policy evaluation in reinforcement learning with human preferences. We consider a two-stage algorithm: (i) We use a fully connected ReLU network to learn the reward function from human choices by neural maximum likelihood estimation, and (ii) we leverage the learned reward to approximate Q-functions by minimizing a Bellman mean square error, with another ReLU neural network as the function approximator. Our theory proves that when state-action space exhibits low-dimensional structures and the underlying reward function enjoys a smooth representation, FQE with human preference data achieves a $\tilde{O}( H K^{-\frac{\alpha}{2\alpha +d}})$ 
finite-sample estimation error, which matches the standard FQE case in which the reward is observable. 

\bibliography{iclr2024_conference}
\bibliographystyle{iclr2024_conference}
\newpage
\appendix
\onecolumn

\section{Proof for Theorem \ref{thm-reward-estimate}}\label{sec:rew-proof}
\begin{proof}
    In this proof, we denote the human labeling policy with $\rho_{b,h}(\cdot\mid (\tilde{s}_h^{(1)}, \tilde{a}_h^{(1)}), (\tilde{s}_h^{(2)}, \tilde{a}_h^{(2)}))$. Recall that  the preference for step $h$ and episode $k$ is chosen among $\{(\tilde{s}_{h,k}^{(1)}, \tilde{a}_{h,k}^{(1)}),(\tilde{s}_{h,k}^{(2)}, \tilde{a}_{h,k}^{(2)}), (s^f,a^f) \}$. In this proof,  we use the notation of $\Upsilon_{h}^k := \big\{(\tilde{s}_{h,k}^{(1)}, \tilde{a}_{h,k}^{(1)}),(\tilde{s}_{h,k}^{(2)}, \tilde{a}_{h,k}^{(2)}), (s^f,a^f)\big\}$. Recall that by Assumption \ref{ass:rew-model} and \ref{ass:rew-iden},  $\Upsilon_h^k$ is sampled by distribution $\eta_h$ for every $k\in[K]$ , and the human preference $y_{h,k} = (s,a)$ is generated according to \begin{align*}
    \rho_{b,h}( s,a \mid \Upsilon_h^k)= \frac{\exp(r_h(s, a))}{\exp(r_h(\tilde{s}_{h,k}^{(1)}, \tilde{a}_{h,k}^{(1)}))+\exp(r_h(\tilde{s}_{h,k}^{(2)}, \tilde{a}_{h,k}^{(2)}))+ \exp(r_h(s^f,a^f))},
\end{align*}
here the state-action pair $(s,a)\in \Upsilon_{h}^k$. 
For all $h\in[H]$, we define a set $\Upsilon_h:= \{(s_h^{(1)}, a_h^{(1)}),(s_h^{(2)}, a_h^{(2)}), (s^f, a^f)\}$, where $(s_h^{(1)}, a_h^{(1)})$ and $(s_h^{(2)}, a_h^{(2)})$ are fixed state-action pairs. Then $\Upsilon_h^k$ can be regarded as the realization of $\Upsilon$ under distribution $\eta_h$. We also define a vector-value function class defined on $\Upsilon_h$, which is parametrized by function  $r\in\cG$:  $$
 \Pi_h(\Upsilon_h) = \bigg\{ \rho \in \R^3 \mid
 \rho_r(s,a\mid \Upsilon_h) = {\exp({r(s,a)})}/{\sum_{(s',a')\in\Upsilon} \exp({r(s',a')})} \text{ for some } r\in\cG
 \bigg\}.$$
Let $\Ncal_{[]}(\Pi_h,\|\cdot\|_\infty,1/n)$ be the smallest $1/n$-upper bracketing of $\Pi_h$. And $|\Ncal_{[]}(\Pi_h,\|\cdot\|_\infty,1/n)| = N_{[]}(\Pi_h,\|\cdot\|_\infty,1/n)$, where $N_{[]}(\Pi_h,\|\cdot\|_\infty,1/n)$ is the  bracketing number of $\Pi_h$. For a given neural network $\cG$ and a reward function $r$, we denote $$
\epsilon_{misspec} := \min_{r'\in \cG} \|r' - r\|_{\infty}, \qquad r_{h,o} = \operatorname{argmin}_{r'\in \cG} \|r' - r\|_{\infty}.
$$
to measure the approximation ability of $\mathcal{G}$ for the $r$.
Our proof for Theorem \ref{thm-reward-estimate} lies in three steps: \begin{itemize}
    \item[(i)] First, we prove that with MLE, we have  \begin{align}
&\E_{(s_h^{(i)}, a_h^{(i)})\sim \eta_h}\big[\|\rho(\cdot\mid \Upsilon_h ) - \rho^{l}_{h}(\cdot\mid \Upsilon_h)\|_1^2\big] \leq \Ocal\bigg( \frac{\log\big(H\cdot N_{[]}(\Pi_h, \|\cdot\|_{\infty}, 1/K)/\delta\big)}{K_{HF}}\bigg) + \epsilon_{misspec}
\end{align}

with probability at least $1-\delta$. Here $\E_{(s_h^{(i)}, a_h^{(i)})\sim \eta_h}[\cdot]$ means that the expectation is taken under the randomness of $(s_h^{(i)}, a_h^{(i)})\sim \eta_h$, $i\in\{1,2\}$, $\rho_{l,h}$ is the choice probability learned by MLE in step $h$, and the $L^1$ norm is taken over all possible state action pairs in $\Upsilon_{h}$. By invoking Lemma \ref{lem:generalization} we get the expectation version. 

\item[(ii)] Second, we prove that with Assumption \ref{ass:rew-model} and \ref{ass:rew-model},  we can convert the previous TV-bound for human choice probability into an estimation error for the reward function $\widehat{r}_h(s,a)$, with an error bound of \begin{align*}
     &\E\bigg[\int_{\Xcal} |r_h(s,a) - (r^l_h(s,a) - r_h^l (s^f,a^f)) |^2 d\eta_h(s,a) \bigg]\\
     &\qquad\qquad\leq O\bigg(\frac{1}{K_{HF}} \log\big(H \cdot N(\cG, \|\cdot\|_\infty, 1/K_{HF})\big) + \epsilon_{misspec}\bigg) .
 \end{align*}
 Recall that $r_h^l$ is the reward function we learned directly from MLE.
 Note that such a bound measures the tradeoff between approximation and generalization: for the network to have better approximation ability, i.e. a smaller $\epsilon_{misspec}$, a larger neural network is necessary, which increases the complexity of the neural network that is measured in term $\log\big(H \cdot N(\cG, \|\cdot\|_\infty, 1/K_{HF})\big)$.
\item[(iii)] Third, we prove that we can handle the tradeoff between approximation and generalization by choosing an appropriate neural network have \eqref{eq:rew-eps} hold. Specifically, under Assumption \ref{ass:rew-smooth}, we can obtain a fast convergence rate for our estimation $\widehat{r}_h$.
\end{itemize}

\paragraph{Step (i).} For notation simplicity, we denote $\rho_{r_{h,o}}$ by $\rho_{o,h}$ and $\rho_{r_{h,l}}$ by $\rho_{l,h}$.
By MLE procedure, we have \begin{align*}
    \frac{1}{K_{HF}}\sum_{k=1}^{K_{HF}}\log\bigg(\frac{\rho_{l,h}(y_{h,k}\mid \Upsilon_h^k)}{\rho_{b,h}(y_{h,k}\mid \Upsilon_h^k)}\bigg) &= \frac{1}{K_{HF}}\sum_{k=1}^{K_{HF}}\log\bigg(\frac{\rho_{l,h}(y_{h,k}\mid \Upsilon_h^k)}{\rho_{o,h}(y_{h,k}\mid \Upsilon_h^k)}\bigg)+\frac{1}{K_{HF}}\sum_{k=1}^{K_{HF}}\log\bigg(\frac{{\rho}_{o,h}(y_{h,k}\mid \Upsilon_h^k)}{\rho_{b,h}(y_{h,k}\mid \Upsilon_h^k)}\bigg)\\
    &\geq \frac{1}{K_{HF}}\sum_{k=1}^{K_{HF}} \log(\rho_{b,h}(y_{h,k}\mid \Upsilon_h^k)) - \frac{1}{K_{HF}}\sum_{k=1}^{K_{HF}} \log({\rho}_{o,h}(y_{h,k}\mid \Upsilon_h^k))\\
    & = \frac{1}{K_{HF}}\sum_{k=1}^{K_{HF}} \bigg\{{r}_h (s_{h,k}, a_{h,k}) - r_{h,o}(s_{h,k},a_{h,k})\\
    &\qquad \qquad + \log\big(\sum_{(s,a)\in \Upsilon_h} \exp(r_{h,o}(s,a))\big) -  \log\big(\sum_{(s,a)\in \Upsilon_h} \exp(r_{h}(s,a))\big)\bigg\}\\
    &\geq  -4\epsilon_{misspec},
\end{align*}
here the first inequality comes from $\widehat{r}_h^l \in \cG$ and $\rho_{l,h} = \argmax_{\rho_r, r\in \cG} \frac{1}{K}\sum_{k=1}^K \log\big({{\rho}_h(y_h^k \mid s_h^k)}\big)$, the second inequality comes from the fact that $\log(\sum_{i=1}^3 \exp(x_i))$ being $1$-Lipschitz for every component. Meanwhile, by Markov's inequality and Boole’s inequality, it holds with probability at least $1-\delta$ that for all $\bar{\rho}\in\Ncal_{[]}(\Pi,\|\cdot\|_\infty,1/n)$, we have \begin{align*}
    \sum_{k=1}^{K_{HF}} \frac{1}{2}\log\bigg(\frac{\bar{\rho}(y_h^k\mid \Upsilon_h^k)}{\rho_{b,h}(y_h^k\mid \Upsilon_h^k)}\bigg) &\leq K_{HF}\log\bigg(\E_{(s_h^{(i)},a_h^{(i)})\sim \eta_h, y_h\sim \rho_{b,h}}\bigg[\exp\bigg(\frac{1}{2}\log\bigg(\frac{\bar{\rho}(y_h\mid \Upsilon)}{\rho_{b,h}(y_h\mid \Upsilon)}\bigg) \bigg)\bigg]\bigg)\\
    &\qquad+\log\bigg(\frac{N_{[]}(\Pi_h,\|\cdot\|_\infty, 1/K_{HF})}{\delta}\bigg),
\end{align*}
specify $\bar{\rho}$ to be the upper bracket of $\rho_{l,h}$,  we have \begin{align*}
    -2K_{HF}\cdot\epsilon_{misspec} &\leq K_{HF}\log\bigg(\E_{(s_h^{(i)},a_h^{(i)})\sim \eta_h, y_h\sim \rho_{b,h}}\bigg[\exp\bigg(\frac{1}{2}\log\bigg(\frac{\bar{\rho}(y_h\mid \Upsilon_h)}{\rho_{b,h}(y_h\mid \Upsilon_h)}\bigg) \bigg)\bigg]\bigg)\\
    &\qquad+\log\bigg(\frac{N_{[]}(\Pi_h,\|\cdot\|_\infty, 1/K_{HF})}{\delta}\bigg)\\
    &\leq K_{HF}\cdot \log\bigg(\E_{(s_h^{(i)},a_h^{(i)})\sim \eta_h, y_h\sim \rho_{b,h}}\bigg[\sqrt{\frac{\bar{\rho}(y_h\mid \Upsilon_h)}{\rho_{b,h}(y_h\mid \Upsilon_h)}}\bigg]\bigg)+\log\bigg(\frac{N_{[]}(\Pi_h,\|\cdot\|_\infty, 1/K_{HF})}{\delta}\bigg)\\
    &= K_{HF}\cdot\log\bigg(\E_{(s_h^{(i)},a_h^{(i)})\sim \eta_h}\bigg[\sum_{y_h\in\Upsilon_h} \sqrt{\bar{\rho}(y_h\mid \Upsilon_h)\cdot \rho_{b,h}(y_h\mid \Upsilon_h)}\bigg] \bigg)\\
    &\qquad+ \log\bigg(\frac{N_{[]}(\Pi_h,\|\cdot\|_\infty,1/K_{HF})}{\delta}\bigg),
\end{align*}
 Utilizing the $\log x \leq x-1$, we have $$
1- \E_{(s_h^{(i)},a_h^{(i)})\sim \eta_h}\bigg[\sum_{y_h \in \Upsilon_h} \sqrt{\bar{\rho}(y_h\mid \Upsilon_h)\cdot \rho_{b,h}(y_h\mid \Upsilon_h)}\bigg] \leq \frac{1}{K_{HF}} \log\bigg(\frac{N_{[]}(\Pi_h,\|\cdot\|_\infty,1/K_{HF})}{\delta}\bigg)+\epsilon_{misspec}.
$$
Therefore we can bound the Hellinger distance between $\rho_{b,h}$ and $\bar{\rho}$,
\begin{align}\label{hell-dist-1}
    h(\bar{\rho},\rho_{b,h}) &= \E_{(s_h^{(i)},a_h^{(i)})\sim \eta_h}\bigg[\sum_{y_h \in \Upsilon_h}\big(\bar{\rho}(a\mid s_h)^{1/2} - \rho_{b,h}(a\mid s_h)^{1/2} \big)^2\bigg]\\
    &\leq 2\E_{(s_h^{(i)},a_h^{(i)})\sim \eta_h}\bigg[1-\sum_{y_h \in \Upsilon_h} \sqrt{\bar{\rho}(a\mid s_h)\cdot \rho_{b,h}(a\mid s_h)}\bigg] + \frac{1}{K_{HF}}\\
    &\leq \frac{2}{K_{HF}}\log\bigg(\frac{N_{[]}(\Pi,\|\cdot\|_\infty,1/K_{HF})}{\delta}\bigg)+\frac{1}{K_{HF}}+\epsilon_{misspec},
\end{align}
here the second inequality comes from the fact that $\bar{\rho}$ is a upper bracketing of $\Pi$. Moreover, it is easy to verify that \begin{align}\label{eq:hell-dis-2}
\E_{(s_h^{(i)},a_h^{(i)})\sim \eta_h}\big[\sum_{y_h \in \Upsilon_h}((\bar{\rho}(a\mid s_h)^{1/2} + \rho_{b,h}(a\mid s_h)^{1/2} ))^2\big] &\leq 2\E_{(s_h^{(i)},a_h^{(i)})\sim \eta_h}\big[\sum_{y_h \in \Upsilon_h}(\bar{\rho}(a\mid s_h) + \rho_{b,h}(a\mid s_h))\big] \\
&\leq \frac{2}{K_{HF}} +4,
\end{align}
where the second inequality comes from the fact that $\bar{\pi}$ is the $1/n$-upper bracket of a probability distribution.
Combining the \eqref{hell-dist-1} and \eqref{eq:hell-dis-2}, by Cauchy-Schwarz inequality, we have \begin{align*}
&\E_{(s_h^{(i)},a_h^{(i)})\sim \eta_h}\big[\|\bar{\rho}(\cdot\mid \Upsilon_h) - \rho_{b,h}(\cdot\mid \Upsilon_h)\|_1^2\big]\\
&\qquad\qquad\leq \frac{15}{K_{HF}}\cdot \log\bigg(\frac{N_{[]}(\Pi,\|\cdot\|_\infty,1/K_{HF})}{\delta}\bigg) + \epsilon_{misspec},
\end{align*}
here the $L_1$ norm is taken under all $(s,a) \in \Upsilon_h$. Meanwhile, \begin{align*}
    &\|\bar{\rho}(\cdot\mid \Upsilon_h) - \rho_{b,h}(\cdot\mid \Upsilon_h)\|_1^2 - \|\rho_{l,h}(\cdot\mid \Upsilon_h) - \rho_{b,h}(\cdot\mid \Upsilon_h)\|_1^2\leq\bigg(4+\frac{1}{K_{HF}}\bigg)\cdot\frac{1}{K_{HF}},
\end{align*}
therefore we have \begin{align*}   &\E_{(s_h^{(i)},a_h^{(i)})\sim \eta_h}\big[\|\rho_{l,h}(\cdot\mid \Upsilon_h) - \rho_{b,h}(\cdot\mid \Upsilon_h)\|_1^2\big]\leq \frac{20}{K_{HF}}\cdot\log\bigg(\frac{N_{[]}(\Pi_h,\|\cdot\|_\infty,1/K_{HF})}{\delta}\bigg)+ \epsilon_{misspec}.
\end{align*}
By Lemma 
Next, we bound $
N_{[]}(\Pi_h,\|\cdot\|_\infty,1/K) $ by  $N(\cG,\|\cdot\|_\infty,1/4K) $.
For all $h\in[H]$, recall the definition  $$
 \Pi_h(\Upsilon_h) = \big\{
 \rho_r(a\mid s) = {\exp({r(s,a)})}/{\sum_{a'\in\Upsilon_h} \exp({r(s,a')})} \text{ for some } r\in\Fcal_h
 \big\}.
 $$
 It is easy to check that $$
 |\rho_r(a\mid s) - \rho_{r'}(a\mid s)| \leq 2\cdot \|r - r'\|_{\infty} , \forall (s,a)\in\Scal\times\Acal.
 $$
 Recall that $N(\Fcal_h,\|\cdot\|_\infty,1/K)$ is the covering number for model class $\Fcal_h$.  We have \begin{equation}\label{eq:cover-brack}
    N_{[]}(\Pi_h,\|\cdot\|_\infty,1/K_{HF})\leq N(\cG_h,\|\cdot\|_\infty,1/4K_{HF})
\end{equation}
always hold for all $h\in[H]$ by Lemma \ref{lem:cover-bracket}. 
Thus we conclude the proof in \textbf{Step (i)}.
\paragraph{Step (ii).} 
Note that by Lemma \ref{lem:generalization}, we can turn into a bound in expectation: \begin{align}\label{eq:expect-bound}
&\E\bigg[\E_{(s_h^{(i)},a_h^{(i)})\sim \eta_h}\big[\|{\rho}_{l,h}(\cdot\mid \Upsilon_h) - \rho_{b,h}(\cdot\mid \Upsilon_h)\|_1^2\big]\bigg] \\
&\qquad\leq O\bigg(\frac{1}{K_{HF}} \log\big(H \cdot N(\cG_h, \|\cdot\|_\infty, 1/K_{HF})\big) + \epsilon_{misspec}\bigg),\nonumber
\end{align}
where the expectation is taken with respect to the data-generating process. Recall that \begin{align*}
    &\rho_{l,h}((s,a)\mid \Upsilon_h) = \frac{\exp(r_h^l (s,a) )}{\sum_{a\in\Upsilon_h} \exp(r_h^l(s,a))},\\
    &\rho_{b,h}((s,a)\mid \Upsilon_h) = \frac{\exp(r_h (s,a) )}{\sum_{a\in\Upsilon_h} \exp(r_h(s,a))},\nonumber
\end{align*} with $r_h(s^f, a^f) = 0$.
By basic algebra, we have \begin{align*}
    \sum_{(s,a)\in \Upsilon_h } |r_h(s,a) - (r^l_h(s,a) - r^l(s^f, a^f)|^2 &= \sum_{(s,a)\in \Upsilon_h } \bigg|\log\bigg(\frac{\rho_{l,h}(s,a \mid \Upsilon_h)}{\rho_{b,h}(s,a \mid \Upsilon_h)}\bigg) - \log\bigg(\frac{\rho_{l,h}(s^f,a^f \mid \Upsilon_h)}{\rho_{b,h}(s^f,a^f \mid \Upsilon_h)}\bigg)\bigg|^2 \\
    & \leq 20 \cdot  \|{\rho}_{l,h}(\cdot\mid \Upsilon_h) - \rho_{b,h}(\cdot\mid \Upsilon_h)\|_1^2
\end{align*}
for any arbitrary $\Upsilon_h$. Combine this with \eqref{eq:expect-bound}, we have \begin{align*}
     \E\bigg[\int_{\Xcal} |r_h(s,a) - (r^l_h(s,a) - r_h^l (s^f, a^f)) |^2 d\eta_h(s,a) \bigg] \leq O\bigg(\frac{1}{K_{HF}} \log\big(H \cdot N(\cG_h, \|\cdot\|_\infty, 1/K_{HF})\big) + \epsilon_{misspec}\bigg) .
 \end{align*}
Recall that in Algorithm \ref{alg-main}, we construct $\widehat{r}_h(s,a)  = r^l_h(s,a) - r_h^l (s^f, a^f)$. Therefore we have \begin{align}\label{eq:r-mse-bound}
    \E\bigg[\int_{\Xcal} |r_h(s,a) - \widehat{r}_h(s,a) |^2 d\eta_h(s,a) \bigg] \leq O\bigg(\frac{1}{K_{HF}} \log\big(H \cdot N(\cG_h, \|\cdot\|_\infty, 1/K_{HF})\big) + \epsilon_{misspec}\bigg) .
\end{align}
\paragraph{Step (iii).} Note that the bound in \ref{eq:r-mse-bound} consists of two terms: the approximation error of the neural network $\cG_h$, and the model class complexity of $\cG_h$. With Assumption \ref{ass:rew-smooth}, we connect the \eqref{eq:r-mse-bound} with the selection of a proper neural network class. Recall that $f\in \Hcal^{\alpha_{HF}}([0,1]^d_{HF})$, and $\psi \in \Hcal^{\alpha_{HF}}$. We consider $\widehat{f}(\widehat{\psi}(x))$, with $\widehat{\phi}: \Mcal \rightarrow [0,1]^{d_{HF}}$. Then the approximation error in the norm  $L^\infty$ has the following decomposition:\begin{align}\label{local-1}
    |\widehat{f}\circ\widehat{\psi}(x)- f\circ\psi(x)| &\leq |\widehat{f}\circ \widehat{\psi}(x) - \widehat{f}\circ\psi(x)|+ |\widehat{f}\circ\psi(x) - f\circ \psi(x)|\\
    &\lesssim \|\widehat{f} - f\|_\infty + \max_{i\in[d^{HF}]}\|\widehat{\psi}_i - \psi_i\|_\infty,
\end{align} 
therefore to bound the approximation error of $r = f\circ \psi$, we only need to approximate $\psi$ and $f$ respectively. By Lemma 2 in \cite{nguyentang2022sample}, for every function $h \in \Hcal^{\alpha_{HF}}$ and bounded in $[0, R]$ and a neural network class $\mathcal{E}(L,p,I,\tau, R)$, if  \begin{align*}
L=O\left(\log \frac{1}{\epsilon}\right)&, p=O\left(\epsilon^{-\frac{d_{HF}}{\alpha_{HF}}}\right), I=O\left(\epsilon^{-\frac{d_{HF}}{\alpha_{HF}}} \log \frac{1}{\epsilon}\right), \\
&\tau=\max \left\{B, 1, \sqrt{d}, \omega^2\right\}, R=1,
\end{align*}
then we have$$
\min_{f \in \mathcal{E}} \|f - h\|_\infty \leq \epsilon.
$$
therefore, by \eqref{local-1}, to guarantee $\min_{g\in\cG_h(\tilde{R}, \tilde{\tau}, \tilde{L},\tilde{p}, \tilde{I})} \|g - r\|_\infty \leq O(\epsilon)$, we only need to guarantee that $\|f - \hat{f}\|_\infty \leq \epsilon$ and $\|\psi_h - \hat{\psi}^i_h\|_\infty\leq \epsilon$ respectively. It suffices to choose \begin{align}\label{loc-2}
        \tilde{L}=O\bigg(&\log \frac{1}{\epsilon}\bigg), \tilde{p} =O\bigg(\epsilon^{-\frac{d_{HF}}{\alpha_{HF}}}\cdot d_{HF}\bigg), \tilde{I} =O\bigg(\epsilon^{-\frac{d_{HF}}{\alpha_{HF}}}\cdot d_{HF}\cdot\log \frac{1}{\epsilon}\bigg),\nonumber\\
    &\tilde{\tau} = \max\{B, 1, \sqrt{d_{HF}}, \omega^2\}, \tilde{R} = 1.
    \end{align}
Note that by Lemma \ref{lem:cover-bound}, this implies a log-covering number of \begin{align}\label{loc-3}
\log N(H\cdot\cG_h, \|\cdot\|_\infty, 1/K_{HF})&\leq O\bigg(\log\bigg(\frac{2 L^2(p B+2) \kappa^L p^{L+1}}{1/K_{HF}}\bigg)^{K_{HF}}\bigg) \\
&\leq O\bigg(\epsilon^{-d_{HF}/\alpha_{HF}} \cdot\log^3\frac{2}{\epsilon}\cdot\frac{\log K_{HF}}{K_{HF}}\bigg).\nonumber
\end{align}
Recall that in \textbf{Step (ii)} we have $$
\E\bigg[\int_\Xcal (r_h(s,a) - \widehat{r}_h(s,a) )^2 d\eta_h(s,a)\bigg] \leq O\bigg(\epsilon + \log N(H\cdot\cG_h, \|\cdot\|_\infty, 1/K)\bigg).
$$
Set $\epsilon = \epsilon_{misspec}$ in \eqref{loc-2} and \eqref{loc-3}, then we have $$
\E\bigg[\int_\Xcal (r_h(s,a) - \widehat{r}_h(s,a) )^2 dq_h^{\pi}\bigg] \leq O\bigg(\epsilon_{misspec} + \epsilon_{misspec}^{-d_{HF}/\alpha_{HF}} \cdot\log^3\frac{2}{\epsilon_{misspec}}\cdot\frac{\log K_{HF}}{K_{HF}}\bigg)
$$
for all $\cG_h(\tilde{R}, \tilde{\tau}, \tilde{L},\tilde{p}, \tilde{I})$ with $h\in[H]$.
Finally, set $\epsilon_{misspec} = \epsilon_0^2$, and
we conclude the proof of Lemma \ref{thm-reward-estimate}.
\end{proof}
\newpage
\section{Proof For Theorem \ref{thm-main-fqe}}\label{sec:proof-main}
\begin{proof}
    In Theorem \ref{thm-main-fqe}, we aim to bound the $L_1$ distance between our estimate $v^{\pi} $ and $ \widehat{v}^{\pi}$. We decompose error recursively for a summation of $H$-in step errors and then change the probability measure. The key is to find a upper bound for $\E_{(s,a)\sim q_h^\pi(s,a)}\bigg[\bigg|\widehat{\Tcal}_h^\pi\widehat{Q}_{h+1}^\pi - \Tcal_h^\pi Q_{h+1}^\pi \bigg|\bigg]$. First, by definition of $\hat{v}^\pi$ and Bellman equation, we have
    \begin{align*}
    \E\big|v^\pi - \widehat{v}^\pi\big| &= \E\bigg[\bigg|\int_{\Xcal} (Q_1^\pi - \widehat{Q}_1^\pi)(s,a)dq_1^\pi(s,a)\bigg|\bigg]\\
    &\leq \E\bigg[\int_{\Xcal} \big|Q_1^\pi - \widehat{Q}_1^\pi\big|(s,a)dq_1^\pi(s,a)\bigg],
\end{align*}
here the expectation is taken with respect to the data generating process, i.e. $(s,a)\sim q_1^\pi$. The first equality comes from the Bellman equation, and the second inequality comes from Jensen's inequality.
Now we analyze how the estimation error would propagate with the horizon length $H$, and prove our result by mathematic induction. For the expected error for $\widehat{Q}_h$, we have \begin{align*}
    &\E\bigg[\int_{\Xcal} \big|Q_h^\pi(s,a) - \widehat{Q}_h^\pi(s,a)\big|\bigg]\\
    &\qquad= \E\bigg[\int_\Xcal \big| \Tcal_h^\pi Q_{h+1}^\pi(s,a) - \widehat{\Tcal}_h^\pi\widehat{Q}_{h+1}^\pi(s,a)\big|dq_h^\pi(s,a)\bigg]\\
    &\qquad\leq \underbrace{\E\bigg[\int_\Xcal \big| \Tcal_h^\pi Q_{h+1}^\pi(s,a) - \Tcal_h^\pi\widehat{Q}_{h+1}^\pi(s,a)\big|dq_h^\pi(s,a)\bigg]}_{\text{(i)}}\\
    &\qquad\qquad+\underbrace{\E\bigg[\int_\Xcal \big| \Tcal_h^\pi \widehat{Q}_{h+1}^\pi(s,a) - \widehat{\Tcal}_h^\pi\widehat{Q}_{h+1}^\pi(s,a)\big|dq_h^\pi(s,a)\bigg]}_{\text{(ii)}},\\
\end{align*}
To bound (i), note that $\Tcal f(s,a) - \Tcal g(s,a) = \E\big[f(s',a') - g(s',a')\mid s,a\big]$, where $(s',a')$ is the succeeding state-action pair after $(s,a)$. Therefore by Jensen's inequality, we have \begin{align*}
    \operatorname{(i)}&\leq \E\bigg[\int_\Xcal \E\big[\big|Q_{h+1}(s',a') - \widehat{Q}_{h+1}(s',a')\big|\mid s,a\big]dq^\pi_h(s,a)\bigg]\\
    &\leq \E\bigg[\int_\Xcal \big|Q_{h+1}(s,a) - \widehat{Q}_{h+1}(s,a)\big|dq^\pi_{h+1}(s,a)\bigg].
\end{align*}
Next, we bound (ii) by the distribution shift captured by $\chisq$ distance. For notation simplicity, we define the surrogate Bellman operator: $$\bar{\Tcal}_h^\pi \widehat{Q}_{h+1}^\pi := \widehat{r}_h + \cP^\pi_h\widehat{Q}_{h+1}^\pi,
$$ 
and we have \begin{align*}
    \operatorname{(ii)}&\leq \E\bigg[\int_\Xcal \big| \widehat{\Tcal}_h^\pi \widehat{Q}_{h+1}^\pi(s,a) - \bar{\Tcal}_h^\pi\widehat{Q}_{h+1}^\pi(s,a)\big|dq_h^\pi(s,a)\bigg] + \E\bigg[\int_\Xcal \big| \Tcal_h^\pi \widehat{Q}_{h+1}^\pi(s,a) - \bar{\Tcal}_h^\pi\widehat{Q}_{h+1}^\pi(s,a)\big|dq_h^\pi(s,a)\bigg] \\
    & = \underbrace{\E\bigg[\int_\Xcal \big| \widehat{\Tcal}_h^\pi \widehat{Q}_{h+1}^\pi(s,a) - \bar{\Tcal}_h^\pi\widehat{Q}_{h+1}^\pi(s,a)\big|dq_h^\pi(s,a)\bigg]}_{\text{(iii)}} + \underbrace{\E\bigg[\int_\Xcal \big| \widehat{r}_h(s,a) - r_h(s,a)\big|dq_h^\pi(s,a)\bigg]}_{\text{(iv)}}.
\end{align*}
We now aim to bound (iii) and (iv) by the distribution shift of $\Dcal$ and $\Dcal^{HF}$ with respect to $\pi$, respectively. First, notice that \begin{align*}
    \operatorname{(iv)}&\leq \E\bigg[\E\bigg[\sqrt{\int_\Xcal \big(r_h(s,a) -  \widehat{r}_h(s,a) \big)^2d\eta_h(s,a)}\cdot \sqrt{1+\chisq(q_h^\pi, \eta_h)}\mid \Dcal_{h+1}^{HF}, \dots,\Dcal_H^{HF}\bigg]\bigg]\nonumber\\
    &\leq \sqrt{1+\chisq(q_h^\pi, \eta_h)}\cdot\sqrt{\E\bigg[\E\bigg[{\int_\Xcal \big(r_h(s,a) -  \widehat{r}_h(s,a) \big)^2d\eta_h(s,a)}\mid \Dcal_{h+1}, \dots,\Dcal_H\bigg]\bigg]}\nonumber\\
    &=\sqrt{1+\chisq(q_h^\pi, \eta_h)}\cdot \sqrt{\E\bigg[{\int_\Xcal \big(r_h(s,a) -  \widehat{r}_h(s,a) \big)^2d\eta_h(s,a)}\bigg]},
\end{align*}
By Theorem \ref{thm-reward-estimate}, we have $$
\E\bigg[{\int_\Xcal \big(r_h(s,a) -  \widehat{r}_h(s,a) \big)^2d\eta_h(s,a)}\bigg] \leq O\bigg(\epsilon_0^2 +\epsilon_0^{-2\tilde{d}/\tilde{\alpha}} \cdot\frac{\log K_{HF}}{K_{HF}}\cdot\log^3\frac{2}{\epsilon_0}\bigg)
$$
for all $\epsilon_0 \in (0,1)$ and $\cG$ defined in Theorem \ref{thm-reward-estimate}. Therefore we have $$
\operatorname{(iv)} \leq \sqrt{1+\chisq(q_h^\pi, \eta_h)} \cdot O\bigg( \epsilon_0^2 +\epsilon_0^{-2\tilde{d}/\tilde{\alpha}} \cdot\frac{\log K_{HF}}{K_{HF}}\cdot\log^3\frac{2}{\epsilon_0}\bigg).
$$

For (iii), we have 
\begin{align}\label{eq:tech1}
    \operatorname{(iii)} &\leq \E\bigg[\E\bigg[\sqrt{\int_\Xcal \big(\bar{\Tcal}_h^\pi \widehat{Q}_{h+1}^\pi(s,a) - \widehat{\Tcal}_h^\pi \widehat{Q}_{h+1}^\pi(s,a) \big)^2dq^{\pi_0}_{h}(s,a)}\cdot \sqrt{1+\chisq(q_h^\pi, q_h^{\pi_0})}\mid \Dcal_{h+1}, \dots,\Dcal_H\bigg]\bigg]\nonumber\\
    &\leq \sqrt{1+\chisq(q_h^\pi, q_h^{\pi_0})}\cdot\sqrt{\E\bigg[\E\bigg[{\int_\Xcal \big(\bar{\Tcal}_h^\pi \widehat{Q}_{h+1}^\pi(s,a) - \widehat{\Tcal}_h^\pi \widehat{Q}_{h+1}^\pi(s,a) \big)^2dq^{\pi_0}_{h}(s,a)}\mid \Dcal_{h+1}, \dots,\Dcal_H\bigg]\bigg]}\nonumber\\
    & = \sqrt{1+\chisq(q_h^\pi, q_h^{\pi_0})}\cdot\sqrt{\E\bigg[{\int_\Acal \big(\bar{\Tcal}_h^\pi \widehat{Q}_{h+1}^\pi(s,a) - \widehat{\Tcal}_h^\pi \widehat{Q}_{h+1}^\pi(s,a) \big)^2dq^{\pi_0}_{h}(s,a)}\bigg]}.
\end{align}
The first inequality comes from Lemma \ref{lem:dis-shift}, and the third equality comes from the definition of conditional expectation. By \eqref{eq:tech1}, we bound the estimation error under the distribution induced by $\pi$ by the distribution shift between $\pi_0$ and $\pi$. Now we are operating on the distribution induced by $\pi_0$, and there's no distribution shift between distribution induced by $\pi_0$ and the data generating distribution.

Now we have the following decomposition:\begin{align*}
    &\E\bigg[\int_{\Xcal}\big(\bar{\Tcal}_h^\pi \widehat{Q}_{h+1}^\pi(s,a) - \widehat{\Tcal}_h^\pi \widehat{Q}_{h+1}^\pi(s,a)\big)^2dq_h^{\pi_0}(s,a)\bigg]\\
    &\qquad=  \E\bigg[\frac{2}{K}\sum_{k=1}^K\big(\bar{\Tcal}_h^\pi \widehat{Q}_{h+1}^\pi(s_{h,k},a_{h,k}) - \widehat{\Tcal}_h^\pi \widehat{Q}_{h+1}^\pi(s_{h,k},a_{h,k})\big)^2\bigg]\\
    &\qquad \qquad + \E\bigg[\int_{\Xcal}\big(\bar{\Tcal}_h^\pi \widehat{Q}_{h+1}^\pi(s,a) - \widehat{\Tcal}_h^\pi \widehat{Q}_{h+1}^\pi(s,a)\big)^2dq_h^{\pi_0}(s,a)\bigg] \\
    &\qquad\qquad- \E\bigg[\frac{2}{K}\sum_{k=1}^K\big(\bar{\Tcal}_h^\pi \widehat{Q}_{h+1}^\pi(s_{h,k},a_{h,k}) - \widehat{\Tcal}_h^\pi \widehat{Q}_{h+1}^\pi(s_{h,k},a_{h,k})\big)^2\bigg]\\
    &\qquad = \epsilon_{approx} + \epsilon_{general},
\end{align*}
where we define the empirical approximation error and generalization error respectively, i.e. \begin{align*}
    &\epsilon_{approx}:=\E\bigg[\frac{2}{K}\sum_{k=1}^K\big(\bar{\Tcal}_h^\pi \widehat{Q}_{h+1}^\pi(s_{h,k},a_{h,k}) - \widehat{\Tcal}_h^\pi \widehat{Q}_{h+1}^\pi(s_{h,k},a_{h,k})\big)^2\bigg],\\
&\epsilon_{general}:=\E\bigg[\int_{\Xcal}\big(\bar{\Tcal}_h^\pi \widehat{Q}_{h+1}^\pi(s,a) - \widehat{\Tcal}_h^\pi \widehat{Q}_{h+1}^\pi(s,a)\big)^2dq_h^{\pi_0}(s,a)\bigg]\\
    &\qquad\qquad\qquad- \E\bigg[\frac{2}{K}\sum_{k=1}^K\big(\bar{\Tcal}_h^\pi \widehat{Q}_{h+1}^\pi(s_{h,k},a_{h,k}) - \widehat{\Tcal}_h^\pi \widehat{Q}_{h+1}^\pi(s_{h,k},a_{h,k})\big)^2\bigg].
\end{align*}
Here both expectation are taken with respect to the dataset generating process. By Lemma \ref{lem:generalization}, we have $$
\epsilon_{general} \leq \frac{104 R^2}{3 K} \log \mathcal{N}\left(\delta / 4 R, \mathcal{F}(R, \kappa, L, p, K),\|\cdot\|_{\infty}\right)+\left(4+\frac{1}{2 R}\right) \delta.
$$

We now only need to bound the empirical approximation error $\epsilon_{approx}$.  Lemma \ref{lem:cover-bound-mse} allows us to have  we have\begin{align}\label{bound-approx}
\epsilon_{approx} &\leq 4 \inf _{f \in \mathcal{F}(R, \mathcal{k}, L, p, I)} \int_{\mathcal{X}}\left(f(s,a)-\bar{\Tcal}_h^\pi \widehat{Q}_h^\pi(s,a)\right)^2 d q_h^{\pi_0}(s,a)\nonumber\\
&\quad\qquad+48 \sigma^2 \frac{\log \mathcal{N}\left(\delta, \mathcal{F}(R, \kappa, L, p, I),\|\cdot\|_{\infty}\right)+2}{K} \nonumber\\
&\quad\qquad+\left(8 \sqrt{6} \sqrt{\frac{\log \mathcal{N}\left(\delta, \mathcal{F}(R, \kappa, L, p, I),\|\cdot\|_{\infty}\right)+2}{K}}+8\right) \sigma \delta,
\end{align}
where $\sigma$ is the variance of the sample point $\widehat{r}_h(s_{h,k},a_{h,k}) + \int_\Acal \widehat{Q}_h(s_{h,k},a)\pi(a\mid s_{h,k})da$  with mean $\bar{\Tcal}_h\widehat{Q}_h(s_h,a_h)$, strictly upper bounded by $HR$. Note that \eqref{bound-approx} is controlled by two terms: the $L^2$ approximation ability of $\Fcal_h$ under distribution $q_h^\pi$, and the logarithm of function class covering number, which represents the function class complexity. We now only need to estimate the term $\inf _{f \in \mathcal{F}(R, \mathcal{k}, L, p, I)} \int_{\mathcal{X}}\left(f(s,a)-\bar{\Tcal}_h^\pi \widehat{Q}_h^\pi(s,a)\right)^2 d q_h^{\pi_0}(s,a)$. Here the only problem is $\widehat{r}_h$ is generated by a neural network and is therefore not Holder in general, which makes it untractable. Nevertheless, by Theorem \ref{thm-reward-estimate}, the difference between $\hat{r}_h$ and the \holder function $r_h$ can be bounded, and we have \begin{align}
    &\inf _{f \in \mathcal{F}(R, \tau, L, p, I)} \int_{\mathcal{X}}\left(f(s,a)-\bar{\Tcal}_h^\pi \widehat{Q}_h^\pi(s,a)\right)^2 d q_h^{\pi_0}(s,a)\nonumber\\
    \leq& 2\bigg\{
    \inf_{f \in \mathcal{F}(R, \tau, L, p, I)} \int_{\mathcal{X}}\left(f(s,a)-r_h(s,a)-\mathcal{P}_h^\pi \widehat{Q}_h^\pi(s,a)\right)^2 d q_h^{\pi_0}(s,a)\nonumber\nonumber\\
    &\qquad\qquad + \int_\Xcal \big(r_h(s,a) - \widehat{r}_h(s,a)\big)^2 dq_h^{\pi_0}(s,a)\bigg\}\nonumber\\
    \leq&2\bigg\{
    \inf_{f \in \mathcal{F}(R, \tau, L, p, I)} \int_{\mathcal{X}}\left(f(s,a)-r_h(s,a)-\mathcal{P}_h^\pi \widehat{Q}_h^\pi(s,a)\right)^2 d q_h^{\pi_0}(s,a)\nonumber\nonumber\\
    &\qquad\qquad+ O\bigg(\epsilon_0^2 +\epsilon_0^{-2d_{HF}/\alpha_{HF}} \cdot\log^3\frac{2}{\epsilon_0}\cdot\frac{\log K_{HF}}{K_{HF}}\bigg)\nonumber,\\
\end{align}
Plugging this back to $\epsilon_{approx}$ and combining with $\epsilon_{approx}$, we have \begin{align*}
    \operatorname{(ii)} &\leq  \bigg\{\inf_{f \in \mathcal{F}(R, \tau, L, p, I)} \int_{\mathcal{X}}\left(f(s,a)-r_h(s,a)-\mathcal{P}_h^\pi \widehat{Q}_h^\pi(s,a)\right)^2 d q_h^{\pi_0}(s,a)\nonumber \\
    &\qquad\qquad+O\bigg(\epsilon_0^2 +\epsilon_0^{-2d_{HF}/\alpha_{HF}} \cdot\log^3\frac{2}{\epsilon_0}\cdot\frac{\log K_{HF}}{K_{HF}}\bigg)\nonumber \\
    &\qquad\qquad + \frac{R^2}{K} \log \mathcal{N}\left(\delta / 4 R, \mathcal{F}(R, \kappa, L, p, K),\|\cdot\|_{\infty}\right)+\left(4+\frac{1}{2 R}\right) \delta\nonumber\\
    &\qquad\qquad+\sigma^2 \frac{\log \mathcal{N}\left(\delta, \mathcal{F}(R, \kappa, L, p, I),\|\cdot\|_{\infty}\right)}{K}\nonumber \\
&\qquad\qquad+\sigma \delta\cdot\sqrt{\frac{\log \mathcal{N}\left(\delta, \mathcal{F}(R, \kappa, L, p, I),\|\cdot\|_{\infty}\right)}{K}} \bigg\}^{1/2}\cdot\sqrt{1+ \chi^2_{\Qcal}(q_h^\pi, q_h^{\pi_0})}\nonumber\\
&\qquad \qquad+  O\bigg(\epsilon_0^2 +\epsilon_0^{-2d_{HF}/\alpha_{HF}} \cdot\log^3\frac{2}{\epsilon_0}\cdot\frac{\log K_{HF}}{K_{HF}}\bigg)\cdot\sqrt{1+\chi^2_{\Qcal}(q_h^\pi, \eta_h)}.
\end{align*}
Combining everything together, we have \begin{align}\label{eq:bias-var}
    &\E\bigg[\int_\Xcal |Q_h^\pi(s,a) - \widehat{Q}_h^\pi (s,a)|dq_h(s,a)\bigg]\nonumber\\
    &\qquad\lesssim\E\bigg[\int_\Xcal |Q_{h+1}^\pi(s,a) - \widehat{Q}_{h+1}^\pi (s,a)|dq_{h+1}(s,a)\bigg]\nonumber\\
    &\qquad\qquad+\bigg\{\inf_{f \in \mathcal{F}(R, \tau, L, p, I)} \int_{\mathcal{X}}\left(f(s,a)-r_h(s,a)-\mathcal{P}_h^\pi \widehat{Q}_h^\pi(s,a)\right)^2 d q_h^{\pi_0}(s,a)\nonumber \\
    &\qquad\qquad+ O\bigg(\epsilon_0^2 +\epsilon_0^{-2d_{HF}/\alpha_{HF}} \cdot\log^3\frac{2}{\epsilon_0}\cdot\frac{\log K_{HF}}{K_{HF}}\bigg)\nonumber \\
    &\qquad\qquad + \frac{R^2}{K} \log \mathcal{N}\left(\delta / 4 R, \mathcal{F}(R, \tau, L, p, K),\|\cdot\|_{\infty}\right)+\left(4+\frac{1}{2 R}\right) \delta\nonumber\\
    &\qquad\qquad+\sigma \delta\cdot\sqrt{\frac{\log \mathcal{N}\left(\delta, \mathcal{F}(R, \tau, L, p, I),\|\cdot\|_{\infty}\right)}{K}}\nonumber\\
    &\qquad\qquad+\sigma^2 \frac{\log \mathcal{N}\left(\delta, \mathcal{F}(R, \tau, L, p, I),\|\cdot\|_{\infty}\right)}{K}\nonumber\bigg\}^{1/2}\cdot \sqrt{1+\chi^2_\Qcal(q_h^\pi,q_h^{\pi_0})}\nonumber\\
    &\qquad \qquad+  O\bigg(\epsilon_0^2 +\epsilon_0^{-2d_{HF}/\alpha_{HF}} \cdot\log^3\frac{2}{\epsilon_0}\cdot\frac{\log K_{HF}}{K_{HF}}\bigg)\cdot\sqrt{1+\chi^2_{\Qcal}(q_h^\pi, \eta_h)}.
\end{align} 
To balance the bias-variance tradeoff, we only need to initialize the size of function class $\Fcal_h$ and $\cG_h$ deliberately. By Lemma 2 in \cite{nguyentang2022sample} and Assumption \ref{ass-low-dim-sa}, in order to yield a function $f$ such that $\|f - \bar{\Tcal}_h^\pi \widehat{Q}_h^\pi \|_\infty \leq \epsilon$, we need \begin{align*}
    L=O\bigg(&\log \frac{1}{\epsilon}\bigg), p =O\bigg(\epsilon^{-\frac{d}{\alpha}}\bigg), I =O\bigg(\epsilon^{-\frac{d}{\alpha}}\cdot \log \frac{1}{\epsilon}\bigg),\\
    &\tau = \max\{B, H, \sqrt{d}, \omega^2\}, R = H,
\end{align*}
recall that $d$ is the intrinsic dimension of $(s,a)$, $B$ is the upper bound of $(s,a)$ in $L_\infty$, $\omega$ is the reach of Riemannian manifold. $O(\cdot)$ hides factors of $\log D, \alpha, d$ and surface area of $\Xcal$. 

By Lemma \ref{lem:cover-bound}, for a ReLU neural network $\Hcal$, we have \begin{align*}
\log \Ncal (\delta, \Hcal(L,p,I,\tau, R)) &= \log\bigg(\frac{2L^2(pB+2)\tau^Lp^{L+1}}{\delta}\bigg)^I\\
&\lesssim \epsilon^{-\frac{d}{\alpha}} \log^3\frac{1}{\epsilon}\log\frac{1}{\delta},
\end{align*}
here $\Hcal \in \{\Fcal, \cG\}$, we hide the constant depending on $\log B, \omega$ and $\log\log n$. Plugging it in, we have \begin{align*}
    &\E\bigg[\int_\Xcal |Q_h^\pi(s,a) - \widehat{Q}_h^\pi (s,a)|dq_h(s,a)\bigg]\nonumber\\
    &\qquad\leq\E\bigg[\int_\Xcal |Q_{h+1}^\pi(s,a) - \widehat{Q}_{h+1}^\pi (s,a)|dq_{h+1}(s,a)\bigg]\nonumber\\
    &\qquad\qquad+\bigg\{O\bigg(\epsilon_0^2 +\epsilon_0^{-2d_{HF}/\alpha_{HF}} \cdot\log^3\frac{2}{\epsilon_0}\cdot\frac{\log K_{HF}}{K_{HF}}\bigg)
    \nonumber \\
    &\qquad\qquad + \frac{H^2}{K} \bigg((\epsilon^{-d/\alpha}+D)\log^3\frac{1}{\epsilon}\log\frac{1}{\delta}\bigg)+\left(4+\frac{1}{2 H}\right) \delta\nonumber\\
    &\qquad\qquad+H \delta\cdot\sqrt{\frac{\epsilon^{-d/\alpha}+\log^3\frac{1}{\epsilon}\log\frac{1}{\delta}}{K}}  \bigg\}^{1/2} \cdot \sqrt{1+\chi^2_\Qcal(q_h^\pi,q_h^{\pi_0})}\nonumber\\
    &\qquad\qquad + O\bigg(\epsilon_0^2 +\epsilon_0^{-2d_{HF}/\alpha_{HF}} \cdot\log^3\frac{2}{\epsilon_0}\cdot\frac{\log K_{HF}}{K_{HF}}\bigg) \cdot \sqrt{1+ \chi^2_\Qcal(q_h^\pi,\eta_h)}.
\end{align*}
Now we need to balance the tradeoff between model complexity and model approximation ability. We choose $\epsilon,\epsilon_0$ to satisfy $\epsilon^2 = \frac{1}{K} \epsilon^{-d/\alpha}$, $\epsilon_0 = \frac{1}{K_{HF}}\epsilon_0^{-2d/\alpha}$, which gives $\epsilon =  K^{-\frac{\alpha}{2\alpha + d}}$ and $\epsilon_0 = K^{-\frac{\alpha_{HF}}{2\alpha_{HF} + 2 d_{HF}}}$. It suffices to pick $\delta = 1/K$. And we conclude that \begin{align*}
        &\E\bigg[\int_\Xcal |Q_h^\pi(s,a) - \widehat{Q}_h^\pi (s,a)|dq_h(s,a)\bigg]\nonumber\\
    &\qquad\leq\E\bigg[\int_\Xcal |Q_{h+1}^\pi(s,a) - \widehat{Q}_{h+1}^\pi (s,a)|dq_{h+1}(s,a)\bigg]\\
    &\qquad\qquad+ c\cdot \sqrt{1+\chi_{\Qcal}^2(q_h^\pi, q_h^{\pi_0})} H \bigg(K^{-\frac{\alpha}{2\alpha +d}}+ K_{HF}^{-\frac{\alpha_{HF}}{2\alpha_{HF}+2d_{HF}}}+\sqrt{\frac{D}{K}}\bigg)\log^{3/2} K\\
    &\qquad\qquad + c' \cdot \sqrt{1+\chi_{\Qcal}^2(q_h^\pi, \eta_h)}\cdot K_{HF}^{-\frac{\alpha_{HF}}{2\alpha_{HF}+2d_{HF}}},
\end{align*}
where constant $c$ depends on $\log D, \log K_{HF},d, \alpha, p,q, B, \omega$ and surface area of $\Xcal$. 
Note that by Lemma 2 in \cite{nguyentang2022sample}, it suffices to choose $\Fcal(R,\tau, L, p, I)$ such that
\begin{align*}
    L = O\bigg( \frac{\alpha}{2\alpha +d}\cdot\log K\bigg)&, p =O\bigg(K^{\frac{d}{2\alpha +d}}\bigg), I = O\bigg(\frac{\alpha}{2\alpha +d}\cdot K^{\frac{d}{2\alpha +d}}\cdot\log K\bigg),\\
        & \tau = \max\{B,1,\sqrt{d},\omega^2\}, R =H.
\end{align*}
By our selection of $\epsilon_0 = K^{-\frac{\alpha_{HF}}{2\alpha_{HF} + 2 d_{HF}}}$, by Theorem \ref{thm-reward-estimate}, it suffices to choose $\cG(\tilde{R}, \tilde{\tau}, \tilde{L} , \tilde{p}, \tilde{I})$ such that \begin{align*}
    \tilde{L}=O\bigg( \frac{\alpha_{HF}}{2\alpha_{HF} +d_{HF}}\cdot\log K_{HF}\bigg)&, \tilde{p}  =O\bigg(K_{HF}^{\frac{d_{HF}}{2\alpha_{HF} +d_{HF}}}\cdot d_{HF}\bigg), \tilde{I} =O\bigg(\frac{\alpha}{2\alpha +d}\cdot K_{HF}^{\frac{d_{HF}}{2\alpha_{HF}+d_{HF}}} \cdot d_{HF}\cdot\log K_{HF}\bigg),\\
    &\tilde{\tau} = \max\{B, 1, \sqrt{d_{HF}}, \omega^2\}, \tilde{R} = 1.
    \end{align*}
By simple induction, we can prove that \begin{align*}
\E\bigg[\int_\Xcal |Q_1^\pi(s,a) - \widehat{Q}_1^\pi (s,a)|dq_h(s,a)\bigg] &\leq \sum_{h=1}^H \sqrt{1+\chi_{\Qcal}^2(q_h^\pi, q_h^{\pi_0})}\cdot H \bigg(K^{-\frac{\alpha}{2\alpha +d}}+ \sqrt{\frac{D}{K}}\bigg)\log^{3/2}K\\
&\qquad+\sum_{h=1}^H \bigg(\sqrt{1+\chi_{\Qcal}^2(q_h^\pi, \eta_h)} +  \sqrt{1+\chi_{\Qcal}^2(q_h^\pi, q_h^{\pi_0})}\bigg)\cdot K_{HF}^{-\frac{d_{HF}}{2\alpha_{HF}+d_{HF}}}\cdot \log K^{3/2}.
\end{align*}
Taking expectation and we conclude the proof.

\end{proof}
\newpage
\section{Auxiliary Lemma}
\begin{lemma}
    For a positive random variable $X$ and constant $c_1, c_2 >0$, if with probability at least $1-\delta$ that $X\leq (c_1+ \log(1/\delta))\cdot c_2$ for all $\delta \in [0,1]$, then we have $\E[X] \leq (1+c_1)\cdot c_2$.
    
\end{lemma}
\begin{proof}
    By condition, we have 
    $$
    \P\bigg(\frac{X}{c_2} - c_1 > t\bigg) \leq e^{-t},
    $$
    invoking $\E[X] = \int_{t\geq 0} \P(X> t)d\mu(t)$, and we have 
    \begin{align*}
        \E\bigg[\frac{X}{c_2} - c_1\bigg] \leq \int_{t\geq 0} e^{-t}dt = 1,
    \end{align*}
    and we conclude the proof.
\end{proof}

\begin{lemma}\label{lem:cover-bracket}
Consider a class $\Fcal$ of functions ${m_\theta : \theta\in \Theta}$ indexed by a parameter $\theta$ in an arbitrary
index set $\Theta$ with a metric $d$. Suppose that the dependence on $\theta$ is Lipschitz in the sense that $$\left|m_{\theta_1}(x)-m_{\theta_2}(x)\right| \leq d\left(\theta_1, \theta_2\right) F(x)$$ for some function $F: \Xcal\rightarrow\R$, for every $\theta_1,\theta_2\in\Theta$ and $x\in\Xcal$. Then, for any norm $\|\cdot\|$, the bracketing
numbers of this class are bounded by the covering numbers:$$
N_{[]}(\Fcal,\|\cdot\|, 2\epsilon\|F\|) \leq N(\Theta,d,\epsilon).
$$
\end{lemma}
\begin{proof}
See Lemma 2.14 in \cite{sen2018gentle} for details.
\end{proof}
\begin{lemma}[Lemma 7 in \cite{chen2022nonparametric}]\label{lem:cover-bound}
    Given $\delta>0$, the $\delta$-covering number of the neural network class $\mathcal{F}(R, \kappa, L, p, K)$ satisfies
$$
\mathcal{N}\left(\delta, \mathcal{F}(R, \kappa, L, p, K),\|\cdot\|_{\infty}\right) \leq\left(\frac{2 L^2(p B+2) \tau^L p^{L+1}}{\delta}\right)^I.
$$
\end{lemma}
\begin{lemma}[Lemma 5 in \cite{chen2022nonparametric}]\label{lem:cover-bound-mse}
Fix the neural network class $\mathcal{F}\left(M, L, J, I, \tau_1, \tau_2, V\right)$.
Let $\mathcal{X}$ be a d-dimensional compact Riemannian manifold that satisfies Assumption \ref{ass-low-dim-sa}. We are given a function $f_0 $. We are also given samples $S_n=\left\{\left(x_i, y_i\right)\right\}_{i=1}^n$, where $x_i$ are i.i.d. sampled from a distribution $\mathcal{P}_x$ on $\mathcal{X}$ and $y_i=$ $f_0\left(x_i\right)+\zeta_i$. $\zeta_i$ 's are i.i.d. sub-Gaussian random noise with variance $\sigma^2$, uncorrelated with $x_i$ 's. If we compute an estimator
$$
\widehat{f}_n=\arg \min _{f \in \mathcal{F}} \frac{1}{n} \sum_{i=1}^n\left(f\left(x_i\right)-y_i\right)^2
$$
with the neural network class $\mathcal{F}=\mathcal{F}\left(M, L, J, I, \tau_1, \tau_2, V\right)$. For any constant $\delta \in(0,2 V)$, we have
\begin{align*}
    \E\bigg[\frac{1}{n}\sum_{i=1}^n\big(\widehat{f}_n(x) - f_0(x)\big)^2\bigg] &\leq 2 \inf _{f \in \mathcal{F}\left(M, L, J, I, \tau_1, \tau_2, V\right)} \int_{\mathcal{X}}\left(f(x)-f_0(x)\right)^2 d \mathcal{P}_x(x)\\
& \qquad+24 \sigma^2 \frac{\log \mathcal{N}\left(\delta, \mathcal{F}\left(M, L, J, I, \tau_1, \tau_2, V\right),\|\cdot\|_{\infty}\right)+2}{n} \\
& \qquad+\left(4 \sqrt{6} \sqrt{\frac{\log \mathcal{N}\left(\delta, \mathcal{F}\left(M, L, J, I, \tau_1, \tau_2, V\right),\|\cdot\|_{\infty}\right)+2}{n}}+4\right) \sigma \delta,
\end{align*}

where $\mathcal{N}\left(\delta, \mathcal{F}\left(M, L, J, I, \tau_1, \tau_2, V\right),\|\cdot\|_{\infty}\right)$ denotes the $\delta$-covering number of $\mathcal{F}\left(M, L, J, I, \tau_1, \tau_2, V\right)$ with respect to the $\ell_{\infty}$ norm, i.e., there exists a discretization of $\mathcal{F}\left(M, L, J, I, \tau_1, \tau_2, V\right)$ into $\mathcal{N}\left(\delta, \mathcal{F}\left(M, L, J, I, \tau_1, \tau_2, V\right),\|\cdot\|_{\infty}\right)$ distinct elements, such that for any $f \in \mathcal{F}$, there is $\bar{f}$ in the discretization satisfying $\|\bar{f}-f\|_{\infty} \leq \epsilon$.    
\end{lemma}

\begin{lemma}\label{lem:dis-shift}
    Given a function class $\mathcal{Q}$ that contains functions mapping from $\mathcal{X}$ to $\mathbb{R}$ and two probability distributions $q_1$ and $q_2$ supported on $\mathcal{X}$, for any $g \in \mathcal{Q}$,
$$
\mathbb{E}_{x \sim q_1}[g(x)] \leq \sqrt{\mathbb{E}_{x \sim q_2}\left[g^2(x)\right]\left(1+\chi_{\mathcal{Q}}^2\left(q_1, q_2\right)\right)} .
$$
\end{lemma}
\begin{proof}
    $$
\begin{aligned}
\mathbb{E}_{x \sim q_1}[g(x)] & =\sqrt{\mathbb{E}_{x \sim q_2}\left[g^2(x)\right] \frac{\mathbb{E}_{x \sim q_1}[g(x)]^2}{\mathbb{E}_{x \sim q_2}\left[g^2(x)\right]}} \\
& \leq \sqrt{\mathbb{E}_{x \sim q_2}\left[g^2(x)\right] \sup _{f \in \mathcal{Q}} \frac{\mathbb{E}_{x \sim q_1}[f(x)]^2}{\mathbb{E}_{x \sim q_2}\left[f^2(x)\right]}} \\
& =\sqrt{\mathbb{E}_{x \sim q_2}\left[g^2(x)\right]\left(1+\chi_{\mathcal{Q}}^2\left(q_1, q_2\right)\right)},
\end{aligned}
$$
where the last step is by the definition of $\chi_{\mathcal{Q}}^2\left(q_1, q_2\right):=\sup _{f \in \mathcal{Q}} \frac{\mathbb{E}_{q_1}[f]^2}{\mathbb{E}_{q_2}\left[f^2\right]}-1$.
\end{proof}
\begin{lemma}[Lemma 6 in \cite{chen2022nonparametric}]\label{lem:generalization}
    For any constant $\delta \in(0,2 R),$ the generalization error $\epsilon_{general}$ satisfies
$$
\epsilon_{general} \leq \frac{104 R^2}{3 K} \log \mathcal{N}\left(\delta / 4 R, \mathcal{F}(R, \kappa, L, p, K),\|\cdot\|_{\infty}\right)+\left(4+\frac{1}{2 R}\right) \delta.
$$
\end{lemma}
\newpage
\section*{Checklist}

 \begin{enumerate}

 \item For all models and algorithms presented, check if you include:
 \begin{enumerate}
   \item A clear description of the mathematical setting, assumptions, algorithm, and/or model. [Yes/No/Not Applicable] \textbf{Yes.}
   \item An analysis of the properties and complexity (time, space, sample size) of any algorithm. [Yes/No/Not Applicable] \textbf{Yes.}
   \item (Optional) Anonymized source code, with specification of all dependencies, including external libraries. [Yes/No/Not Applicable] \textbf{N/A.}
 \end{enumerate}

 \item For any theoretical claim, check if you include:
 \begin{enumerate}
   \item Statements of the full set of assumptions of all theoretical results. [Yes/No/Not Applicable] \textbf{Yes.}
   \item Complete proofs of all theoretical results. [Yes/No/Not Applicable] \textbf{Yes.}
   \item Clear explanations of any assumptions. [Yes/No/Not Applicable]     \textbf{Yes.}
 \end{enumerate}

 \item For all figures and tables that present empirical results, check if you include:
 \begin{enumerate}
   \item The code, data, and instructions needed to reproduce the main experimental results (either in the supplemental material or as a URL). [Yes/No/Not Applicable] \textbf{N/A.}
   \item All the training details (e.g., data splits, hyperparameters, how they were chosen). [Yes/No/Not Applicable] \textbf{N/A.}
         \item A clear definition of the specific measure or statistics and error bars (e.g., with respect to the random seed after running experiments multiple times). [Yes/No/Not Applicable] \textbf{Yes.}
         \item A description of the computing infrastructure used. (e.g., type of GPUs, internal cluster, or cloud provider). [Yes/No/Not Applicable] \textbf{N/A.}
 \end{enumerate}
 \item If you are using existing assets (e.g., code, data, models) or curating/releasing new assets, check if you include:
 \begin{enumerate}
   \item Citations of the creator If your work uses existing assets. [Yes/No/Not Applicable] \textbf{N/A.}
   \item The license information of the assets, if applicable. [Yes/No/Not Applicable] \textbf{N/A.}
   \item New assets either in the supplemental material or as a URL, if applicable. [Yes/No/Not Applicable] \textbf{N/A.}
   \item Information about consent from data providers/curators. [Yes/No/Not Applicable] \textbf{N/A.}
   \item Discussion of sensible content if applicable, e.g., personally identifiable information or offensive content. [Yes/No/Not Applicable] \textbf{N/A.}
 \end{enumerate}

 \item If you used crowdsourcing or conducted research with human subjects, check if you include:
 \begin{enumerate}
   \item The full text of instructions given to participants and screenshots. [Yes/No/Not Applicable] \textbf{N/A.}
   \item Descriptions of potential participant risks, with links to Institutional Review Board (IRB) approvals if applicable. [Yes/No/Not Applicable] \textbf{N/A.}
   \item The estimated hourly wage paid to participants and the total amount spent on participant compensation. [Yes/No/Not Applicable] \textbf{N/A.}
 \end{enumerate}

 \end{enumerate}

\end{document}